\documentclass[11pt]{article}

\date{}

\usepackage[utf8]{inputenc} 
\usepackage[T1]{fontenc}    
\usepackage{hyperref}       
\usepackage{url}            
\usepackage{booktabs}       
\usepackage{amsfonts}       
\usepackage{nicefrac}       
\usepackage{microtype}      
\usepackage{xspace}
\usepackage{natbib}
\usepackage{comment}
\usepackage{amsmath}
\usepackage{amsthm}
\usepackage{graphicx}
\usepackage{rotating}

\usepackage{amssymb}
\usepackage{autobreak}
\usepackage{mathtools}
\usepackage{xcolor}
\usepackage{bbm}
\usepackage{algorithm}
\usepackage{algorithmic}

\makeatletter

\newtheorem{definition}{Definition}
\usepackage{xcolor}

\title{Autonomous Passage Planning for a Polar Vessel}

\author{Jonathan D. Smith \footnote{British Antarctic Survey, Cambridge}, Samuel Hall \footnotemark[1], George Coombs \footnotemark[1], James Byrne \footnotemark[1], Michael A.S. Thorne \footnotemark[1],\\ J. Alexander Brearley \footnotemark[1],  Derek Long  \footnote{King's College, London}, Michael Meredith \footnotemark[1], Maria Fox \footnotemark[1]. \thanks{This paper is non-peer reviewed and is under review in ‘Cold Regions Science and Technology’.}}

\begin{document}

\maketitle

\renewcommand{\abstractname}{Abstract}
\begin{abstract}
We introduce a method for long-distance maritime route planning in polar regions, taking into account complex changing environmental conditions. The method allows the construction of optimised routes, describing the three main stages of the process: discrete modelling of the environmental conditions using a non-uniform mesh, the construction of mesh-optimal paths, and path smoothing. In order to account for different vehicle properties we construct a series of data driven functions that can be applied to the environmental mesh to determine the speed limitations and fuel requirements for a given vessel and mesh cell, representing these quantities graphically and geospatially. In describing our results, we demonstrate an example use case for route planning for the  polar research ship the RRS Sir David Attenborough (SDA), accounting for ice-performance characteristics and validating the spatial-temporal route construction in the region of the Weddell Sea, Antarctica. We demonstrate the versatility of this route construction method by demonstrating that routes change depending on the seasonal sea ice variability, differences in the route-planning objective functions used, and the presence of other environmental conditions such as currents. To demonstrate the generality of our approach, we present examples in the Arctic Ocean and the Baltic Sea. The techniques outlined in this manuscript are generic and can therefore be applied to vessels with different characteristics. Our approach can have considerable utility beyond just a single vessel planning procedure, and we outline how this workflow is applicable to a wider community, e.g. commercial and passenger shipping.
\end{abstract}
\newpage
\renewcommand{\abstractname}{Plain Language Summary}
\begin{abstract}
We introduce a method that helps the captain of a vessel by generating suggested navigational routes in polar regions with changing sea ice conditions and ocean currents. The resulting system works similarly to an in-car navigation system, by presenting best routes annotated with metrics such as travel time and fuel usage. Ocean navigation poses a number of difficulties not encountered in the in-car navigation problem: in ocean environments weather conditions are always changing and can be very disruptive, impacting significantly on route structure and metrics. Furthermore, although shipping lanes are present in busy shipping areas, the open ocean has no fixed road system. Our route-planning approach therefore relies on the dynamic construction of a digital representation of the ocean and environmental conditions, in enough detail to support high quality and timely automated decision-making. Our method is parameterised by ship information such as its typical fuel usage under different ocean conditions. We use the example of the British Antarctic Survey vessel the RRS Sir David Attenborough (SDA). We demonstrate some use cases for our method, including planning routes in years with different sea ice conditions; in the Arctic ocean; and under different quality objectives (such as minimising the fuel cost or travel time implied by a journey).
\end{abstract}

\newpage
\section{Introduction}

Master navigators are highly skilled and can plot safe and efficient routes for ships through complex environmental conditions. However, many regions of the polar oceans are either poorly charted or feature highly dynamic conditions\footnote{https://gcaptain.com/safely-navigating-polar-regions/}. The task of efficient route planning involves consideration of many alternatives in both space and time, and the task of finding optimal paths under changing conditions quickly exceeds the capabilities of human navigators. Providing a toolkit that ship captains can use to help generate and explore alternatives is essential to support decision-making. Such a toolkit can minimise transit time, fuel usage and carbon emissions, and give a detailed explanation of how the changing conditions might affect viable routes. In order to achieve these goals the toolkit must be able to rapidly ingest, combine and interpret multiple datasets and use these data to generate high-integrity digital representations of the world and to generate and explain high quality routes between user-defined waypoints.

\begin{figure}[h]
    \centering
    \includegraphics[width=1.0\textwidth,keepaspectratio]{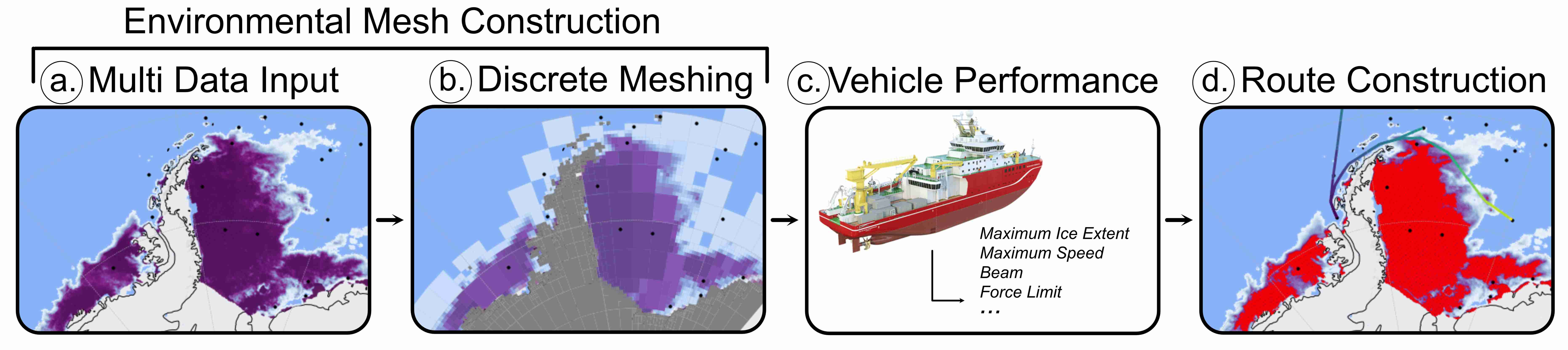}
    \caption{Flow diagram demonstrating the key processing stages of the route-planner toolkit discussed in the following sections}
    \label{fig:FlowDiagram}
\end{figure}

In this manuscript we present an automated route-planning method for use by a vessel operating in polar regions. We build on the prior work developed for underwater vehicle long-distance route planning described in \cite{Fox2021}. We start by generating a meshed model of the environment that is able to ingest multiple different datasets. The mesh is constructed as a quad-tree (\cite{quadtrees}). The sizes of the cells in the mesh are  dependent on a series of functional cell-splitting conditions, leading to a non-uniform mesh with greatest detail in the areas of highest environmental complexity  (Figure \ref{fig:FlowDiagram}a and b). Once the environmental mesh is constructed we apply a series of vessel-dependent functions, computing the vessel speed and fuel usage constraints implied by the mesh properties (e.g. sea-ice concentration, Figure \ref{fig:FlowDiagram}c) in each cell. These functions are used to enrich the cells of the mesh with vehicle-specific information. On this basis we construct mesh-dependent routes for the vessel that satisfy constraints on the performance of the vessel in ice and under other conditions. We finally apply a path-smoothing method to the mesh-based routes, reducing their dependence on the underlying mesh (Figure \ref{fig:FlowDiagram} c) and usually significantly shortening them. This two-stage process of path-generation and path-smoothing efficiently generates routes that follow standard navigation solutions in open water and optimise vessel performance in the face of currents and in and around areas dominated by sea ice.  While we have focused on navigation in and around polar ice, our methods are also applicable to shipping in other conditions (e.g.: commercial shipping) where route-planning must be responsive to changing local and weather conditions.

\section{Background}
\label{sec:background}
There is a large body of work on automated methods for solving the navigation problem for ships and other vehicles in different ocean settings (\cite{Zermelo31,Bijlsma1975,Kotovirta2009,Sen2015,Walther2016,Topaj2019,Lehtola2019,Li2020,Fox2021,Mishra2021}). The most common methods use mesh-based approximations of the environment (\cite{Kotovirta2009,Sen2015,Lehtola2019,Fox2021,Mishra2021}). \cite{Fox2021} introduce a non-uniform meshing strategy, where cells are finest in areas of the greatest complexity. Mesh-based methods discretise the environment into a graph of locations and then use optimisation methods to find paths from the start node to the destination. Approaches to the problem of finding shortest paths in meshes use heuristic search (\cite{Hart1968}) or greedy methods (\cite{Sen2015,Lehtola2019,Fox2021,Mishra2021}) such as Dijkstra's algorithm (\cite{Dijkstra59}). In a mesh-based representation, edges can be weighted to model environmental impacts on vessel performance, such as the presence of strong surface currents, wind or ice (\cite{Mishra2021}). Alternatively, cells can be augmented with performance data that affects route efficiency, as is done here and by~\cite{Lehtola2019}.

 A shortcoming of mesh-based approaches is typically that all paths follow the predefined mesh structure, between vertices that are the centres and corners of the cells. This results in rectilinear paths, consisting of straight horizontal and vertical lines between cell centres, which are often far from optimal depending on the mesh resolution. Further increasing mesh resolution to better approximate optimal solutions is often prohibitively computationally expensive. A further difficulty arises when environmental constraints are time-dependent. These cannot easily be combined into edge weights, because they change over time whilst the edge weights are constant. The 2D mesh used by Dijkstra's algorithm must be extended to a third dimension to model temporal variability, which introduces extra complexity. This complexity is avoided by~\cite{Fox2021} and ~\cite{Lehtola2019}, by introducing time-dependent meshes and a mechanism for passing between layers in the mesh as time moves along.

An alternative, cell-free, modelling approach is to use the principle of wavefront expansion (\cite{Petres2007},\cite{Bijlsma1975}) to identify the points in a non-homogeneous space that can be reached in equal time (or energy expenditure) from the location of the vehicle. The approach avoids discretisation of the space at the outset, but faces similar problems of efficiency and optimality of solutions that depend on the number of points identified on each wavefront. Furthermore, the need for a temporal dimension must still be addressed when conditions are time-dependent.

In~\cite{Mishra2021}, the ice and wind resistance encountered under different conditions are computed using a mathematical model parameterised by vessel-specific information, and precompiled as travel times into the weights connecting nodes in a uniform mesh. Dijkstra's algorithm is then used to compute the mesh-optimal path. Whilst the ice and wind resistance models used are sophisticated, enabling performance in both level and brash ice to be taken into account, the benefits of this sophistication may be outweighed by the simplicity of the route-planning approach. The routes shown reveal typical weaknesses of the mesh-based approach: a highly rectilinear structure which would be infeasible to navigate and are much longer (and require much more fuel) than could be achieved in the absence of the mesh constraints.

\cite{Kotovirta2009} plan routes in a mesh constructed using a lossy compression technique. To address dependence on the underlying mesh, Powell's conjugate direction method (\cite{Powell1964}) is used to search for a route between two points taking into account the impact of different ice conditions on speed. The result is a path that is not constrained by mesh artifacts so is not restricted to the construction of rectilinear paths. A weakness of the approach is that the curvature of the Earth is not taken into account, and the number of waypoints needed for route construction must be determined in advance. The distance to be travelled must be split into segments to keep the overall computational cost realistic. The approach therefore constructs long routes as chains of segments and are hence not globally optimal. Long paths comprising great circle arcs cannot be constructed without modelling of curvature and the ability to choose both number and placement of waypoints dynamically. 

\cite{Lehtola2019} describe an approach to navigation in ice that navigates an icebreaker in a variety of challenging conditions, including those where other marine traffic is present. The path-finding approach used combines a graph-based optimal path-finding method with a post-processing step to remove some of the resulting mesh artifacts from the paths and improve their geodesic validity. This relaxation monotonically improves the cost of the path by successively removing points. When the cost can no longer be improved, the edges between pairs of points are replaced by geodesic curves. The resulting paths have smoothed out some of the mesh effects and respect, to some extent, the curvature of the Earth. A shortcoming of the second part of this approach is that when the cost function has been optimised the resulting path may be far from fully smoothed because of dependence on grid points that could not be removed without worsening the path cost. This is especially the case when the path navigates around ice or land. A further weakness of this approach is that the order of removal of the points is fixed, but removal of different subsets could lead to better paths: a search over alternatives would be prohibitively expensive. 

The navigation problem, which is to find the fastest path for a vehicle navigating in a vector field on a plane, was first formalised by~\cite{Zermelo31}. He offered as an example the problem of navigating a ship in wind. In a conference in 1929 he also proposed a version of the same problem with an airship moving in a 3-dimensional vector field. Zermelo and other mathematicians approached the problem primarily through the calculus of variations. Certain cases are solvable analytically (for example, a vector field of constant magnitude and direction is easily solvable) and navigation on surfaces of revolution, such as a globe or an ellipsoid, have been explored, with some simpler cases also being solvable analytically. In most cases, the problem can only be solved numerically. 

\section{Methods and Data}
\label{sec:methods}
 The following datasets were used for the construction of the environmental mesh used in the work described here. The Southern Ocean current vector field is a 6-year surface current mean, averaged over the years 2005-2010 inclusive, provided by the Southern Ocean State Estimate (SOSE) (\cite{Mazloff2010})\footnote{\url{https://climatedataguide.ucar.edu/climate-data/southern-ocean-state-estimate-sose}}. The Southern Ocean Sea Ice Concentrations (SICs) are obtained from the AMSR-2 (Advanced Microwave Scanning Radiometer) dataset\footnote{\url{https://www.earthdata.nasa.gov/learn/find-data/near-real-time/amsr2}}, for a given year (we used 2014, 2019 and 2022 for the experiments reported in Section~\ref{sec-results}). The depth measurements used were provided by the GEBCO-2022 data set~\footnote{\url{https://www.gebco.net/data_and_products/gridded_bathymetry_data/}}. GEBCO is the global terrain model for ocean and land. We use this data in preference to the Southern Ocean portion (IBCSO) because it supports application in any area of the global oceans. For the Arctic Ocean we used the AMSR-2 Northern Hemisphere data included in the dataset used for the Southern Ocean. We zeroed currents for the Arctic Ocean in the example we studied. For the Baltic sea ice concentration data we use the Copernicus gridded sea ice time series dataset~\footnote{\url{https://doi.org/10.48670/moi-00131}} for March 2011, and for the Baltic Sea currents we use the Copernicus Baltic Sea Physics Reanalysis~\footnote{\url{https://doi.org/10.48670/moi-00013}}.  We are not restricted to the use of any of these datasets: any geo-spatially located data with a defined coordinate system can be reprojected to EPSG 4326 and be ingested by our mesh construction process. 
 
 Wind speed and direction, although having a potentially significant impact on efficiency of transit, is not yet included but is one of the topics of our current work.
 
 The current vector field is not analytic, but is described extensionally at discrete sample points, and the sea ice extent is modelled as a scalar field which is also not described analytically. Finally, the land (any points of less than a chosen threshold depth) presents obstacles that are, once again, not described analytically and present discontinuities in the underlying space. Because we do not in fact have a continuous model, our problem cannot be approached directly using the techniques that have been considered for solving variants of Zermelo's problem.

Instead, we follow a three-stage approach. Firstly, we support a non-uniform meshing so that cells can be refined where there is high variance in the environmental conditions. We model the temporal dimension by discretising time into different time periods and recomputing the meshing in each time period. 

Secondly, we generate mesh-based route plans using Dijkstra's algorithm. We address the limitations of Dijkstra's algorithm, noted above, in three ways.  Locally, the problem of finding the fastest path between two adjacent cells involves coupling two instances of Zermelo's navigation problem each in a constant magnitude and constant direction field (the current) specific to the cell. We address this problem using Newton's method to find the optimal crossing point on the adjoining edge between two adjacent cells, in the presence of currents and ice concentrations. Using Newton's method mitigates the effect of the mesh by allowing the traversal between cells to pass through vertices that are neither centres nor corners of cells. Dijkstra's algorithm is then used to find routes that optimise some user-defined objective function such as travel time. 

Thirdly, having generated mesh-optimal paths using Dijkstra's algorithm, we further mitigate the effect of the mesh by smoothing the paths to remove visits to artifacts of the mesh and to better approximate realistic traverses over the curved surface of the Earth.

 The novelties of our approach, particularly with respect to \cite{Kotovirta2009, Lehtola2019} and \cite{Mishra2021}, who have tackled a very similar problem, are: the use of a non-uniform mesh constructed from a variety of different datasets, the use of Newton's method within the construction of Dijkstra paths, and the post-processing of paths to smooth them, using a method that removes {\em  all} of the mesh artifacts whilst following the Dijkstra path as closely as possible. The combination of these features leads to routes that combine the best characteristics of both the discrete and the continuous methods, whilst having the great benefit of being highly computationally efficient to generate.

\section{Environmental Mesh Construction} \label{sec:mesh}
In this section we will give a brief background to the construction of a discretised environmental mesh representation originally outlined in  \cite{Fox2021}. In this work, the environmental conditions are discretised to construct a high integrity model of a physical environment, using multiple heterogeneous datasets consisting of observational and modelled data. A set of hierarchical splitting conditions are applied to dynamically represent the environmental data as a non-uniform quadtree mesh (\cite{quadtrees}), consisting of a series of independent cells. The different datasets are each independently aggregated inside each cell to reduce the number of data points, making the procedure more computationally efficient for downstream applications like route planning, discussed later in this article. Outlined below, the mesh construction can be separated into three main stages: processing input data, environmental mesh splitting and neighbourhood graph construction.

\subsection{Processing Input Data}
The environment conditions can be represented as a series of distinctly different datasets, all with differing projections and input datatypes (e.g. Scalar, Vector). For the route planner described here, we use information about ocean currents, temporally varying SIC and bathymetry, but want to create a input structure that is insensitive to possible future data formats. This is achieved by defining a series of loader functions that are dependent on the input dataset. These loader functions act to extract the time-period and spatial domain of interest, applying a reprojection to a WGS-84 (EPSG: 4326) geographic reference frame. The transformed data are passed to the construction of the mesh given in the next section.

\subsection{Environmental Mesh Splitting}
A regular meshed representation of the environmental conditions allows for the combination of different datasets which may not be resolved on the same regular mesh. However, care must be taken to define a mesh resolution that enables resolution of the small scale variability in all the datasets, whilst ensuring that the mesh resolution never exceeds that of the dataset and without incurring an unacceptable computation cost. Further improvement can be achieved by defining a dynamic variable sized mesh, where discrete meshing is refined only in areas with large differences, further improving the computational cost.  \cite{Fox2021}  generated a variable size mesh for the environmental conditions, sub-dividing mesh cells with large variance in dataset values into four small cells, based on a specified splitting condition. We follow the same approach. First, the cells in the mesh are scaled uniformly by the latitudes of their centres. This results in rectangular cells, with cells at more southerly latitudes having smaller width. Once a cell is split, the data in the cell is then reallocated amongst the subdivisions and the process of subdivision applied recursively to each of the quarters until a user defined maximum split depth is achieved, or a minimum number of data points in a given cell is reached. These splitting conditions differ according to the feature being modelled (sea ice, depth etc) and are applied according to a hierarchical process to generate the regional environmental mesh as shown in Figure \ref{fig:RegionalMesh}a. In this figure, each cell has a unique index identifier (e.g. $i_5$ as shown in Figure \ref{fig:RegionalMesh}c), cell centre coordinate ($cx$,$cy$), where $cx$ is the longitude of the centre and $cy$ is the latitude of the centre. The width of a cell at the equator is $2w$. The latitude corrected cell width is then $2w\cdot cos(cy)$. 

\subsection{Neighbourhood Graph}
\begin{figure}
    \centering
    \includegraphics[width=0.9\textwidth,keepaspectratio]{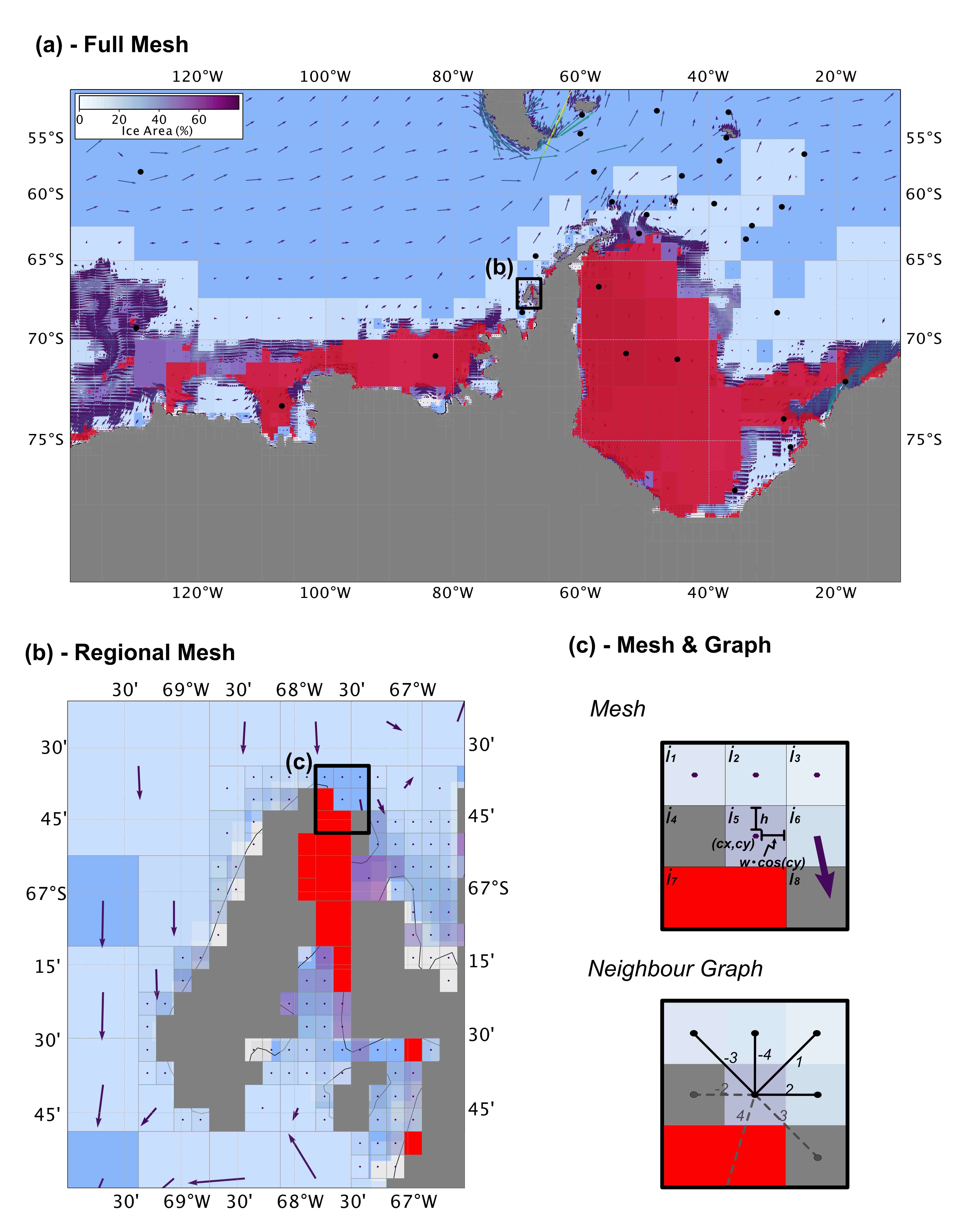}
    \caption{Example of a regional mesh construction. (a) represents the whole region mesh using SIC from 1st Jan 2022. (b) represents a  zoom in. (c) represents the mesh and graph information from a further zoomed in section of the mesh.}
    \label{fig:RegionalMesh}
\end{figure}

Once a regional mesh is generated, the nearest neighbours must be determined for each cell to define the neighbourhood subgraph, with the connections between all cells defining the neighbourhood graph. In order to rapidly identify the directions in which neighbouring cells are adjacent to a given cell, we identify 8 different cases, four of which are diagonals (1 is the case in which travel is heading North East, 3 is the case in which travel is heading South East, -1 is the South West direction and -3 is the North West direction) and the others are orthogonals: 2 is eastward, -2 is westward, 4 is southward and -4 is northward. These arrangements can be seen in Figures \ref{fig:DijkstraInfo}c and~\ref{fig:RegionalMesh}c. 

The neighbourhood graph is pruned so that it only contains the cell connections that can actually be traversed depending on vehicle parameters (eg: maximum SIC tolerance) and environmental conditions (e.g. average SIC, proportion of the cell that is land, and so on). In the example shown in Figure \ref{fig:RegionalMesh}c, these conditions lead to the removal of the neighbours in the westward, southward and south-eastward directions.

\subsection{Performance-based Mesh Augmentation}\label{sec-vesselperf}
In order to plan routes for a given vessel under a given set of environmental conditions we need to construct a model of how its performance will be influenced by the environment conditions in play, and provide the route planner with information about the vessel that will constrain decision-making. In the following sections we describe how we model the effect of sea ice on a generic vessel, how we would subsequently adjust the speed of the vessel to account for this effect and how we can then model the amount of fuel that will be consumed while traversing the cells comprising the planned route.

\subsubsection{Ice Resistance}\label{sec-ice}

When planning routes for a vessel travelling in polar waters one of the key factors to account for is how the vessel in question performs in sea ice conditions. There are certain ice conditions which will be too challenging for the vessel to traverse, depending on the ice breaking capabilities of the ship, and this fact is accounted for by setting an average ice concentration threshold. If any cell in the mesh has an average ice concentration value above this threshold it will be treated as inaccessible by the route planner. The value of this threshold will vary according to the polar class of the vessel in question. 

However, even for ice-infested areas that are considered navigable it is still essential to understand how the ice will influence the performance of the vessel. In order to model the slowing effect of sea ice on a vessel we have implemented the ice resistance model described in~\cite{Colbourne2000}, the mathematical expression of which is shown in Eq.~\ref{eqn:iceres}. This model allows us to determine the resistance force, $R$, experienced by a vessel, taking into account the ice concentration, $C_i$, thickness, $h_i$, and density, $\rho_i$, as well as the beam, $B$, and speed, $V$, of the vessel. Despite recent advances in the Arctic (\cite{Sallila19}), Antarctic sea ice thickness cannot be directly inferred from satellite observations (\cite{Liao22}), so when considering the Southern Ocean we use the quantised approximations based on {\em in situ} measurements described in~\cite{Worby2008}. Whilst these are very coarse approximations they are shown to be realistic by comparison with the few directly sampled data points that we could obtain\footnote{A small number of direct measurements were taken during the ice trials conducted for the RRS Sir David Attenborough in January 2022 (unpublished data)}  and will be improved when observational datasets become available.

\begin{equation}
\label{eqn:iceres}
R = 0.5\, \mathcal{C} \rho_i B h_i V^{2} C_i^{n}
\end{equation}

The factor $\mathcal{C}$ here is the ice force coefficient as defined in Eq.~\ref{eqn:icecoeff}. The model parameters $k_c$, $b$ and $n$ are constants dependent on hull shape and are given in Supplementary Table~\ref{tab:iceparams}.\\

\begin{equation}
\label{eqn:icecoeff}
\mathcal{C} = k_c \mathit{Fr}^{b}
\end{equation}

 $\mathit{Fr}$ is the ice Froude number, as defined in Eq.~\ref{eqn:icefroude}, determined from the ice concentration and thickness as well as the vessel speed and the acceleration due to gravity, $g$.\\
\begin{equation}
\label{eqn:icefroude}
\mathit{Fr} = \frac{V}{\sqrt{gh_iC_i}}
\end{equation} \\

Examples showing how this ice resistance force varies with changing ice concentration for a specific vessel are discussed in Section~\ref{sec-results} but the approach described here is generic and could be applied to any vessel capable of navigating in ice. Using this model the value of the ice resistance force can be calculated for each cell in our discrete mesh and then pre-compiled within that structure for use in calculating other derived quantities. A similar approach of pre-compilation of performance constraints into the mesh is adopted by both ~\cite{Mishra2021} and~\cite{Lehtola2019}. \cite{Mishra2021} captures all such information in the travel time between cells, whilst ~\cite{Lehtola2019} models speed constraints on the vessel within the cells. Our approach goes further in also pre-compiling the fuel requirements implied by the conditions in each cell of the mesh, allowing us to choose routes that minimise fuel use rather than travel time.

\begin{figure}[h]
    \centering
    \includegraphics[width=0.9\textwidth,keepaspectratio]{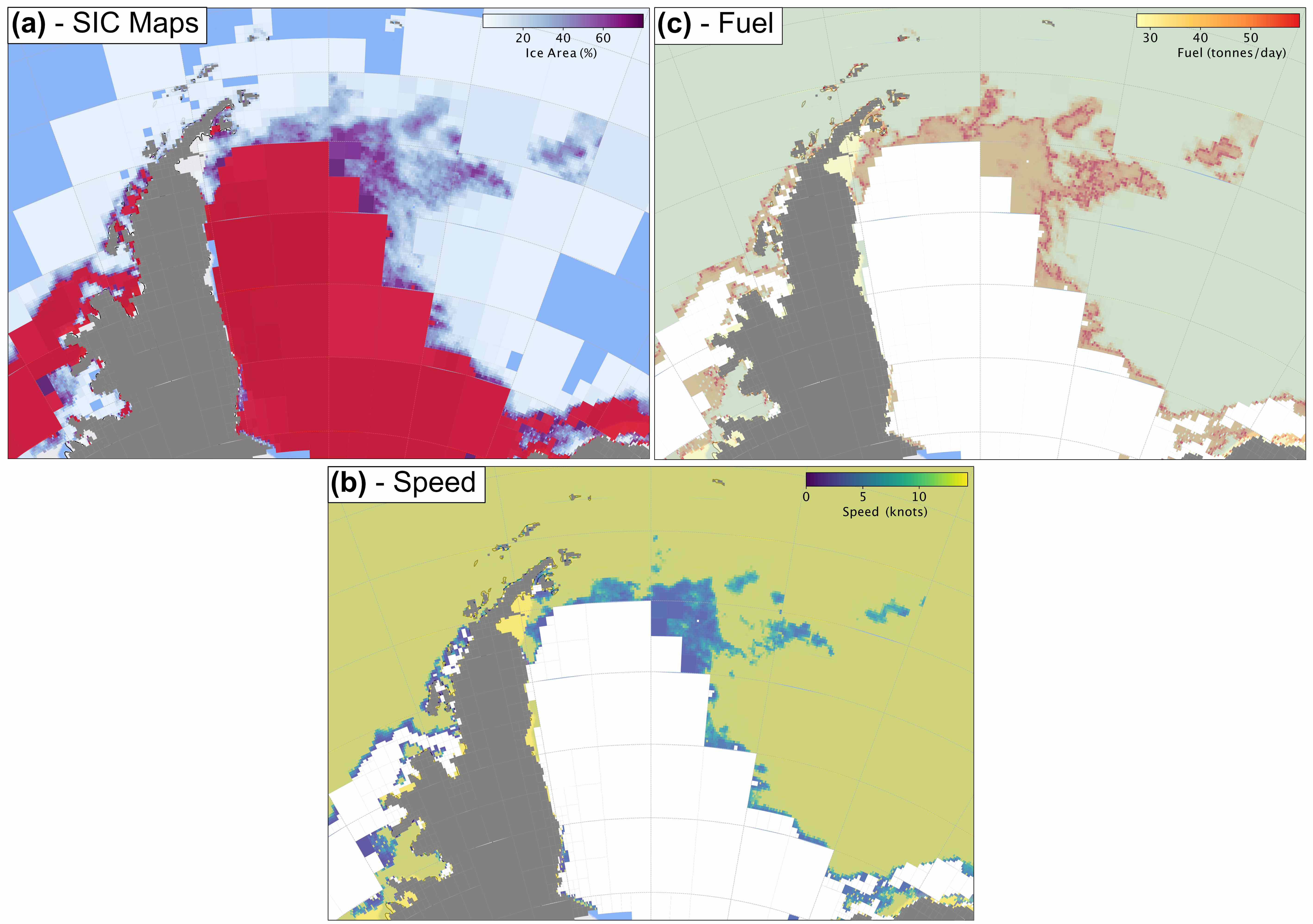}
    \caption{(a) an example sea ice concentration in the Southern Ocean, with red indicating the average SIC over 80\%. (b) The fuel map, with extreme ice shown in white, and brown-scale regions showing the cells with highest (reddest) fuel requirements. (c) The speed map, with ice shown in white and the blue-scale indicating the areas where ice resistance has the highest (bluest) impact on speed.} 
    \label{fig:fuelmap}
\end{figure}

An example of a mesh with the pre-compiled speed and fuel requirements is shown in Figure~\ref{fig:fuelmap}. An interesting consequence of our model is that the fringe of the ice implies a higher fuel requirement than is encountered inside the low-concentration ice. This is because ice resistance is increasing, but has not yet noticeably slowed the ship, so the ship requires more fuel to maintain its speed. This combination of effects is shown in Figure~\ref{fig:iceshipinfo}, in the range between about 40 and 50\% ice concentration, where resistance is increasing sharply but the vessel has not yet started to slow significantly.

\subsubsection{Speed Adjustment}\label{sec-speed}

The maximum speed of the vessel in open water is a configurable parameter of the route planner and this is the default speed that will be assumed for the vessel when planning a route to optimise travel time. However, travelling at this maximum speed will not always be possible. For example, when the vessel encounters significant ice resistance it will be forced to travel more slowly. To account for this and adjust the speed of the vessel to a more realistic value we apply a simple algorithm based on the modelled ice resistance force to calculate the maximum speed in a given cell based on the ice conditions.

This speed adjustment is performed by first determining a maximum resistance force that the vessel can safely handle, checking if a given set of ice conditions produce a resistance force above this threshold and subsequently calculating a new safe speed, if necessary. The force limit, $R_L$ is calculated to be the value of Equation~\ref{eqn:iceres} in the harshest ice conditions a vessel is expected to traverse and at the maximum speed at which it can navigate in those conditions. The force limit is therefore determined by the polar class of the vessel in question. We then calculate the ice resistance force under these conditions at the relevant speed with Eq.~\ref{eqn:iceres} and set this as the maximum speed. The appropriate speed under other conditions can then be calculated from the rearranged form of Eq.~\ref{eqn:iceres} given in Eq.~\ref{eqn:speedfromres}. This relies on the environmental information being available within the mesh as well as the pre-computed value of the ice resistance being stored within each cell.

\begin{equation}
\label{eqn:speedfromres}
V^{2 + b} = \frac{2 R_{L}}{k_c \rho_i B h_i C_i^n (g h_i C_i)^{-\frac{b}{2}}}
\end{equation} \\

\subsubsection{Fuel Consumption} \label{sec-fuel}

Once we have determined the ice resistance force experienced by the vessel in each cell of the meshed environment, and thus the suitable transit speed, it then becomes interesting to examine how much fuel would be required by the vessel to cross that cell as part of a route. We have a developed a simplified fuel consumption model that decomposes the fuel consumption of the vessel into two parts, one determined by the fuel required to operate the vessel at a given speed in open water and the other by the additional fuel required to overcome the calculated ice resistance force. The value for each of these factors will be pre-computed and stored within the discrete data structure of our environment model, in a way similar to that discussed by~\cite{Lehtola2019}. The route planner can then query these values to calculate the fuel consumption rate within a given cell and use that information for optimisation purposes. As the ice resistance force depends directly on the speed of the vessel and the vessel speed is adjusted according to the resistance force encountered this model results in a coupling of the two fuel consumption components when determining fuel optimal routes.

\section{Route Construction} \label{sec:routeplanning}
In this section we will describe our route construction approach, which uses the methods introduced in Section~\ref{sec:methods}. The distances between the cells in the accessibility graph are initialised using Newton's method to optimise the crossing points between adjacent pairs of cells in the presence of currents and ice concentrations. Dijkstra's algorithm is then used to find the optimal routes between pairs of waypoints. These routes are subject to the restrictions on travel speed, fuel use and the effects of ice that have been compiled into the mesh, and are optimal with respect to the constraints imposed by the mesh. They are referred to as {\em mesh-optimal}. This stage of our route-planning method follows the approach described in~\cite{Fox2021}. In the remainder of this paper, we refer to the routes generated by the mesh-based method as {\em Dijkstra paths}. 

An optimal path in open water will in fact always follow a path approximating a great circle arc (subject to the effects of currents). Dijkstra paths are mesh-based approximations of these arcs that are constrained to visit cell centres and corners. The centres of the cells are artifacts of our model and, in large cells,  excursions to the centres can cost hundreds of kilometres (and hence tons of fuel). We therefore proceed to a second stage of route-planning to {\em smooth} the routes to remove visits to the centres and unnecessary visits to the corners of cells. The resulting routes follow close approximations of great circle arcs whenever possible. The smoothing process uses the Dijkstra path as a guide, which greatly improves the efficiency of the construction of smooth routes. The smoothing method we present is an advance over that described by~\cite{Lehtola2019} because it is not constrained to simply bypass a subset of the grid points on the Dijkstra path, but is allowed to freely re-select the crossing points between cells chosen by the Dijkstra path, according to a physical model of the curvature of the Earth. Our approach allows departure from the Dijkstra path with the addition of new cells when our physical model of the curvature suggests that the best smoothed path goes beyond the constraints modelled by the mesh abstraction.


\subsection{Dijkstra Paths - Mesh-Based Route Construction}

\begin{definition} 
\label{defn-adj} 
Two cells, A and B, are {\em adjacent} if neither is blocked and the straight line connecting the centre of A to the centre of B passes through no other cell.
\end{definition}
The mesh abstraction represents the region as a graph, 
\begin{equation}
G = \langle V,E\rangle
\end{equation} 
where $V$ is the set of cell centre points and $E$ is the set of edges connecting adjacent pairs in $V$. 

\begin{figure}
    \centering
    \includegraphics[width=1.0\textwidth,keepaspectratio]{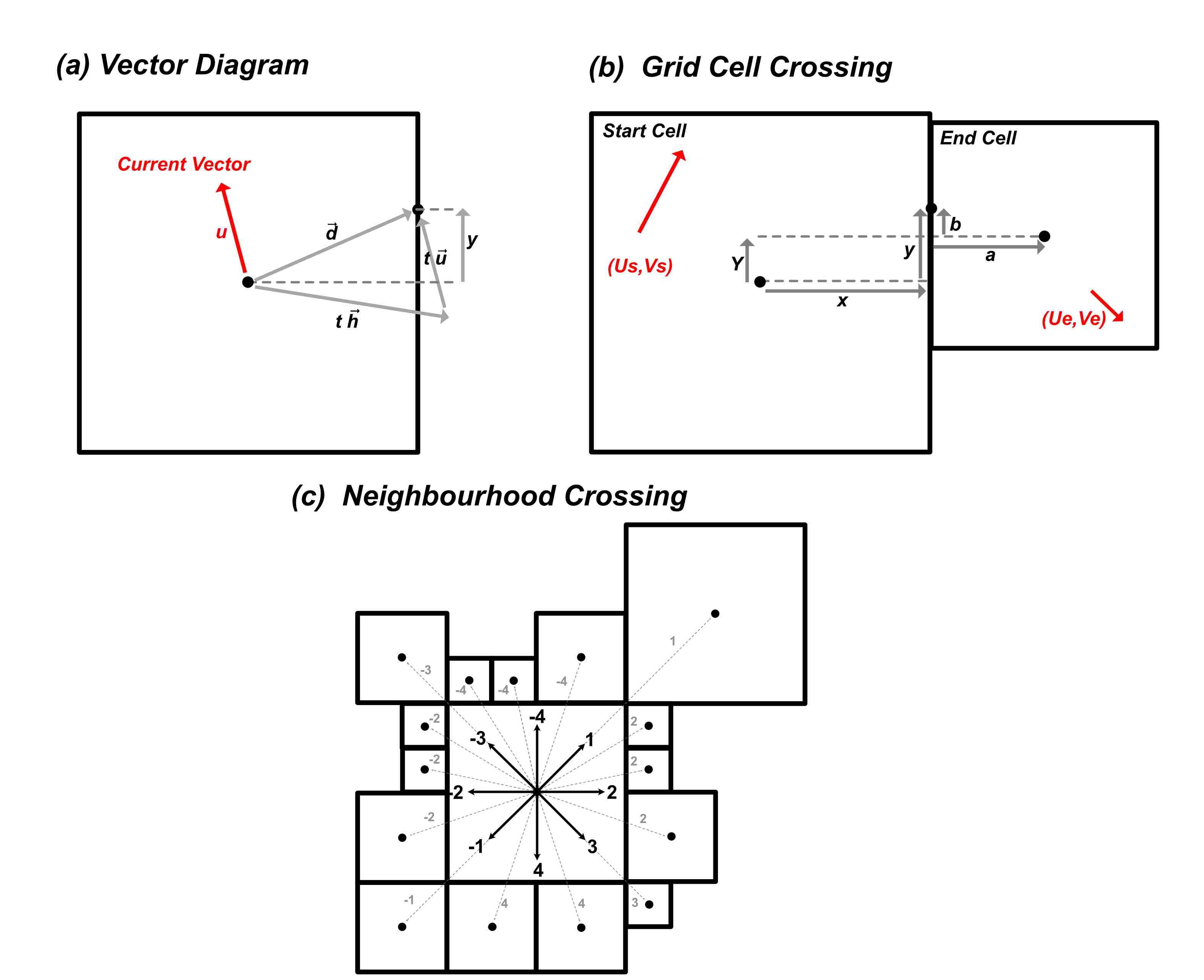}
    \caption{Dijkstra Path construction. (a) represents the vector diagram showing how the time taken to travel through a cell is calculated. (b) represents the finding of the optimal $y$ value crossing crossing point between an adjacent cell pair. The cell widths are scaled by the cosine of their latitude ($\theta$) in order to approximate the Earths curvature. (c) represents the crossing point case definitions, a value between -4 and 4 dependent on the direction.}
    \label{fig:DijkstraInfo}
\end{figure}

In the initial route-planning process a route between waypoints is found by determining the route through the mesh that minimises a user defined objective function. The route between two adjacent cells is found by optimising travel in the cell array representation of the vector field (Figure \ref{fig:DijkstraInfo}a). When the adjacent pair is diagonally arranged, the optimal route between them is initialised to be the straight line between their centres through the corner point connecting the two cells. The non-diagonal cases to be considered are adjacent arrangements where the cell being travelled to is at the left, right, above or below the cell being travelled from. All these arrangements can be normalised into an {\em adjacent cell pair} where the task is to go from the left centre to the right centre, taking into account the currents in the two cells. This requires rotation of the arrangement and consequent interpretation of the $x$ and $y$ axes and the currents.  After normalisation, the route between the adjacent cells goes through an intermediate point on the edge between them, chosen to optimise the travel time between their centre points. This is of course not always the horizontal line, because of the currents present in the cells. The normalised adjacent cell pair arrangement is shown in Figure \ref{fig:DijkstraInfo}b.

The adjacent cell pair has the special property that it is context-free, since it does not matter where the left cell is entered or where the right cell is exited, as these decisions will not affect the selection of the crossing point between the centres.

\begin{figure}
    \centering
    \includegraphics[width=1\textwidth,keepaspectratio]{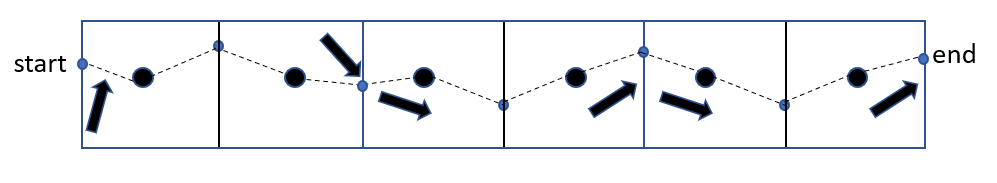}
    \caption{A route is a sequence of connected adjacent cell pairs constituting a directed graph (the edges will have different crossing points in each direction, according to the currents).}
    \label{fig:aroute}
\end{figure}

Having constructed the adjacent cell pairs, we find Dijkstra paths between given start and end waypoints. Dijkstra paths are sequences of connected adjacent cell pairs that minimise a given objective function under the constraints imposed by the mesh. The path structure contextualises the adjacent cell pairs by specifying their entry and exit points. Whilst the objective function used by the Dijkstra method can vary, the resulting paths are constrained by the fact that the crossing points in each of the component adjacent cell pairs are selected by minimising the travel time between the cells as detailed in section ~\ref{sec:newtonsplitcells}. An abtstracted example is shown in Figure~\ref{fig:aroute}. 

\subsubsection{Crossing point optimisation} \label{sec:newtonsplitcells}
The following describes how the crossing point is chosen between two adjacent cells in the initial route-planning process. The method receives parameters that have been scaled by the cosines of the latitudes of the two cells, but does not make any further allowance for curvature. We therefore use an equirectangular approximation in which cells are treated as having a flat surface and a constant width throughout the cell. This allows us to compute the shortest distance across a cell using Pythagoras' theorem as a simple approximation of the distance across the curved surface of the Earth.

In a constant current, the shortest route between two points, ignoring curvature effects, will be the straight line between them. The effect of the current is to slow the vehicle down on this route, as it will have to choose a heading that counters the current. To
achieve the straight line route the vehicle must maintain a constant net velocity along that line, which will
be the sum of the current velocity and the velocity the vehicle is maintaining relative to the water. In our model, the speed of the vehicle in a cell is dependent on the conditions pertaining in that cell.

If the
vehicle crosses the central boundary between cells at a selected point, $yval$ (measured relative to the central crossing point),
then the time, t, it takes to travel from the centre of the left-hand cell to $yval$ (or
from $yval$ to the centre of the right-hand cell) satisfies:
\begin{equation}
\label{eqn:eqn3}
t (\vec{u} + \vec{h}) = \vec{d} 
\end{equation}
where $\vec{u}$ is the current vector in the cell, $\vec{h}$ is the velocity of the vehicle and $\vec{d}$ is the vector from the centre of the cell to the crossing point on the boundary. The cell-dependent speed of the vehicle is constant on this path. 

We proceed as follows. First we find the solution for $t$ as a function of the crossing point $yval$. Then we minimise this function to obtain the $yval$ that optimises the travel time in the cell.
We use Newton's method to optimise the choice of $yval$. Newton's method finds successively better approximations to the roots of a continuous function, iterating until a convergence condition is achieved. The single variable function case is defined as follows:

\begin{equation}
\label{basicnewton}
    x_{n+1} = x_n - \frac{f(x_n)}{f'(x_n)}
\end{equation}

where $f'(x)$ denotes the derivative of $f(x)$.

The function, $f$, that we are minimising is the sum of the two travel times, from the centre of the first cell to the best $yval$ on the edge between the two cells, and from that $yval$ to the centre of the second cell.  These two travel times we denote $tt_l(yval)$ (the time to travel to the $yval$ from the centre of the left cell) and $tt_r(yval)$ (the time to travel to the centre of the right cell from the $yval$). We denote the speed of the vehicle in the left cell as $s_l$ and its speed in the right cell as $s_r$.

Rearranging Equation~\ref{eqn:eqn3}, the travel time in the left cell is derived as follows.

\begin{equation}
\label{eqn:eqn4}
t \vec{h} = \vec{d} - t \vec{u}
\end{equation}
Considering the left cell, in which the speed of the vehicle is $s_l$, the vector from the centre to the crossing point is $d_1$ and the current vector is $u_1$, the square of Equation~\ref{eqn:eqn4} is: 
\begin{equation}
\label{eqn:eqn5}
s_l^2t^2 = |u_1|^2t^2 - 2D_1t + |d_1|^2
\end{equation}
where $D_1 = \vec{u_1}.\vec{d_1}$. This equation can be solved for $t$: 
\begin{equation}
\label{eqn:eqn6}
t =\frac{ (\sqrt{D_1^2 + |\vec{d_1}|^2(s_l^2 - |\vec{u_1}|^2)} - D_1}{s_l^2 - |\vec{u_1}|^2} 
\end{equation}
with a special case when $s_l^2 = |\vec{u_1}|^2$ which is when the speed of the current matches the speed of the vehicle in the left cell. In this case, $t = \frac{|\vec{d_1}|^2}{D_1}$, being undefined when $D_1 = 0$ (the case in which the current is orthogonal to the intended direction of travel). The case when $D_1$ is negative is degenerate, with the vehicle not fast enough to overcome the current in the intended direction. The solution for $t$ defines the function $tt_l(yval)$ for a given current vector, cell size and speed. The relevant vectors are shown in Figure \ref{fig:DijkstraInfo}a. The function $tt_r(yval)$ is defined similarly, using $s_r$, $u_2$, $D_2$ and $d_2$.

Our function, $f(yval)$ is then the sum of the two travel times, $tt_l(yval) + tt_r(yval)$, which we want to minimise by choosing the best $yval$ possible. This is found by iterating over equation~\ref{newton} until a convergence condition is reached. The ticks indicate the first and second derivatives and the subscripted $r$ and $l$ indicate whether travel is through the left or right cell respectively. 
\begin{equation} 
\label{newton}
yval_{k+1} =  yval_k - \frac{tt_l'(yval_k)+tt_r'(yval_k)}{tt_l''(yval_k)+tt_r''(yval_k)}
\end{equation}


The speed of the vehicle within a cell is dependent on the ice conditions that are present. In Section~\ref{sec-vesselperf} we introduce methods for calculating the speed of the vessel in a cell as a function of the ice concentration being encountered in that cell.

The following derivation echoes that given in~\cite{Fox2021}. First we find the solution for $t$ as a function of the crossing point $yval$. Then we minimise this function to obtain the $yval$ that optimises the travel-time in the cell. The variable $y$ stands for the optimal crossing point value $yval$.


Figure \ref{fig:DijkstraInfo}b refers to the following quantities. $t_1$ and $t_2$ are the travel times in the left and right cells respectively, based on $s_l$ and $s_r$, the speeds of the vehicle in the left and right cells. $(u_1,v_1)$ and $(u_2,v_2)$ are the current vector components in the left and right cells respectively. $Y$ is the vertical separation of the two centre points. $x$ is half of the width of the left cell, and $a$ is half of the width of the right cell. As noted above, $y$ is $yval$ itself. $b$ is the vertical distance travelled in the right hand cell. 

The distances travelled in the two cells are calculated as follows.
\begin{equation}
\label{littled1}
d_1^2 = x^2 + y^2
\end{equation}
\begin{equation}
\label{littled2}
d_2^2 = a^2 + b^2 = a^2 + (Y-y)^2
\end{equation}

The following terms are helpful in simplifying the derivations below. We use the shorthand $\partial$ to mean $\frac{d}{dy}$. 
\begin{equation}
\label{c1}
C_1 = s_l^2 - u_1^2 - v_1^2
\end{equation}
\begin{equation}
\label{c2}
C_2 = s_r^2 - u_2^2 - v_2^2
\end{equation}
\begin{equation}
\label{bigd1}
D_1 = u_1x + v_1y
\end{equation}
\begin{equation}
\partial{D_1} = v_1
\end{equation}
\begin{equation}
\label{bigd2}
D_2 = au_2 + (Y-y)v_2
\end{equation}
\begin{equation}
\partial{D_2} = -v_2
\end{equation}
\begin{equation}
\label{x1}
X_1 = \sqrt{D_1^2 + C_1d_1^2}
\end{equation}
\begin{equation}
\label{x2}
X_2 = \sqrt{D_2^2 + C_2d_2^2}
\end{equation}
Equations~\ref{c1} to~\ref{x2} name the left and right cell components of the expressions in equation~\ref{eqn:eqn6}. By pattern matching, equation~\ref{eqn:eqn6} can be specialised to refer to each of the left and right cells, and abbreviated as shown in equations~\ref{t1a} and~\ref{t2a}. These expressions are the travel time functions, $tt_l(y)$ and $tt_r(y)$ respectively.
\begin{equation}
\label{t1a}
    t_1 = \frac{X_1 - D_1}{C_1}
\end{equation}
\begin{equation}
\label{t2a}
    t_2 = \frac{X_2 - D_2}{C_2}
\end{equation}
Differentiation of Equation~\ref{eqn:eqn5} in the left cell, yields:
\begin{equation}
\label{eqn:t1diff1}
\partial{t_1}2C_1t_1+\partial{t_1}2D_1+2v_1t_1 - 2y = 0
\end{equation}
Factorising gives:
\begin{equation}
\label{eqn:t1diff}
\partial{t_1}(C_1t_1+D_1) = y-t_1v_1
\end{equation}
and by a similar derivation for the right cell:
\begin{equation}
\label{eqn:t2diff}
\partial{t_2}(C_2t_2+D_2) = y-Y+t_2v_2
\end{equation}
The minimum travel time is achieved when $\partial{t_1} + \partial{t_2} = 0$. By equation~\ref{t1a} we have that: 

\begin{equation}
C_1t_1 = X_1 - D_1
\end{equation}
so
\begin{equation}
\label{newx1}
X_1 = C_1t_1 + D_1
\end{equation}
and, similarly:
\begin{equation}
\label{newx2}
X_2 = C_2t_2 + D_2
\end{equation}
The derivatives of $X_1$ and $X_2$ are:
\begin{equation}
\partial{X_1} = \frac{D_1v_1 + C_1y}{X_1}
\end{equation}
and 
\begin{equation}
\partial{X_2} = \frac{-D_2v_2 - C_2(Y-y)}{X_2}
\end{equation}
Using~\ref{newx1} with equation~\ref{eqn:t1diff} we have that
\begin{equation}
\partial{t_1} = \frac{y - t_1v_1}{X_1}
\end{equation}
and, by similar reasoning from equation~\ref{t2a}:
\begin{equation}
\partial{t_2} = \frac{y - Y +  t_2v_2}{X_2}
\end{equation}
To obtain the minimum travel time, we require that the derivative of $\partial t_1 + \partial t_2$ is zero:
\begin{equation}
\frac{y - Y + t_2v_2}{X_2} + \frac{y - t_1v_1}{X_1} = 0
\end{equation}
Rewriting this expression using equations~\ref{t1a} and~\ref{t2a}, we define the function $F(y)$:
\begin{equation}
F(y) =  X_2(y - \frac{v_1(X_1 - D_1)}{C_1})  + X_1(y - Y + \frac{v_2(X_2 - D_2)}{C_2}) 
\end{equation}

It can be seen that this equation corresponds to $F(y) = X_1X_2 (tt_l'(y) + tt_r'(y)) $, which is zero whenever $f(y) = \partial t_1 + \partial t_2$ is zero. The derivative of $F(y)$ is:

\begin{equation}
\begin{array}{ll}
\partial{F(y)} = (X_1 + X_2) + y(\partial{X_1} + \partial{X_2})\\
\quad \quad - \frac{v_1}{C_1}( \partial{X_2}(X_1 - D_1) + X_2(\partial{X_1} - \partial{D_1}))\\
\quad \quad + \frac{v_2}{C_2} (\partial{X_1}(X_2 - D_2) + X_1(\partial{X_2} - \partial{D_2}))\\
\quad \quad - Y\partial{X_1}
\end{array}
\end{equation}
 
Using the definitions of $F(y)$ and $\partial{F(y)}$, we can apply Newton's method to arrive at a value of $y$ that makes $F(y) = 0$. 

The differences in vehicle speed between the two cells is due to differences in the ice resistance or other environmental factors encountered in the two cells, as discussed in Section~\ref{sec-vesselperf}. It can arise that there is a marked difference in speed as a vehicle crosses from an ice-free cell into one that contains a significant proportion of ice, resulting in a speed discontinuity at the boundary between cells. In our model, this is similar to a refraction effect because the vehicle must travel at the speeds defined in the two cells. This is a mesh artifact that is mitigated by mesh refinement, since the finer-grained the mesh, the more accurately the physical boundary is captured by the mesh. Furthermore, the effect would be reduced if the ship were able to choose a speed within the limits set by the cells. This is a topic of our future work.

\subsection{Route Smoothing} \label{sec:newtoncurve}
Once the Dijkstra paths have been constructed, they are smoothed in order to remove dependency on the artefacts of the underlying mesh. Avoiding the need to pass through the centres of the cells allows routes to travel further south, often saving substantial travel-time.

When smoothing a route between two waypoints, $a$ and $b$, the Dijkstra path is used to guide the process, as it provides the mesh-optimal sequence of adjacent cell pairs connecting $a$ and $b$. The smoothing process smooths the adjacent cell pairs in the order in which they appear in the Dijkstra path. The task of smoothing the passage between the cells is to pass between the entry and exit points connecting the adjacent cell pairs without visiting the centres of the cells. Smoothing therefore entails choosing the optimal crossing point between the cells when travelling between these entry and exit points rather than between the cell centres. This requires a second invocation of Newton's method, as the optimal crossing point between the entry and exit points will not be the same as the optimal crossing point between the cell centres chosen during Dijkstra path construction. 

On this second invocation, we introduce a more sophisticated approximation of the curvature of the earth that models the change in width {\em within} a cell by introducing a latitude correction. This was not needed in the use of Newton's method during Dijkstra path construction because all paths are forced to go between the centres of the two cells. This requirement dominates any sensitivity to latitude changes within a cell, so the fixed cell-width assumption is adequate in this case. It is no longer adequate when paths are not constrained in this way.

\begin{figure}
    \centering
    \includegraphics[width=1\textwidth,keepaspectratio]{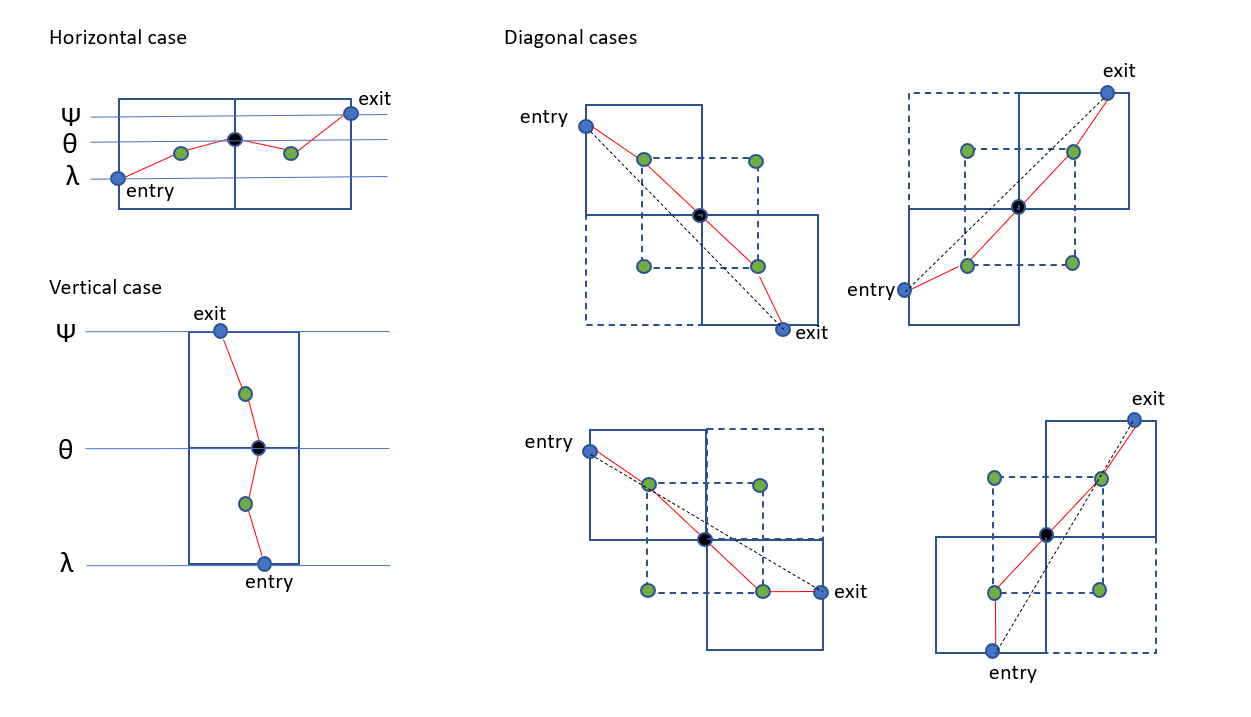}
    \caption{The different cases, shown prior to smoothing, that must be considered for smoothing a route. The entry and exit points are shown in blue, and the chosen crossing point between the centres in black. In the horizontal and vertical cases, adjacent pairs in the horizontal and vertical alignments are shown, contextualised by their entry and exit points. The latitudes of the entry, exit and crossing points are shown. In the diagonal cases, the crossing point is always on the connecting corner. The cells that might be introduced when smoothing the diagonal cases are shown.}
    \label{fig:smoothingcases}
\end{figure}

\begin{figure}
    \centering
    \includegraphics[width=0.8\textwidth,keepaspectratio]{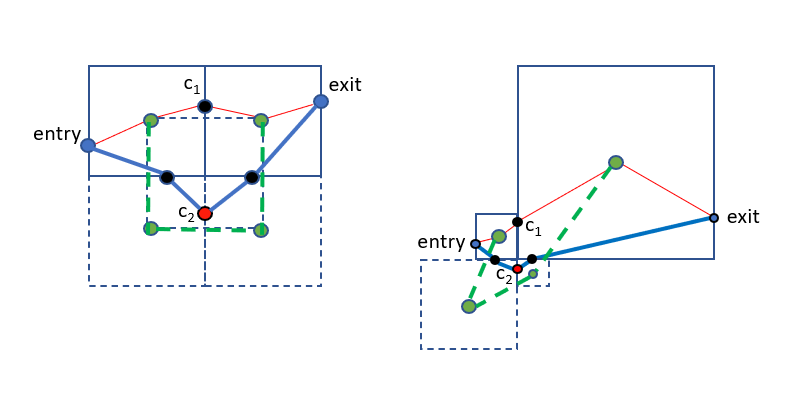}
    \caption{The horseshoe pattern for smoothing a horizontal adjacent cell pair. On the left, the case for two equal-sized cells. On the right, a more general case in which the cell sizes in the original pair differ. The original crossing point in the Dijkstra path is $c_1$. The edges of the Dijkstra path are shown in red. The new crossing point chosen during smoothing is $c_2$. In both cases, $c_2$  drops further south than $c_2$, due to the improved curvature approximation used in the smoothing process, requiring the introduction of a horseshoe pattern. The green nodes and dotted lines are not part of the smoothed path, but indicate the three new adjacent cell pairs that comprise the horseshoe pattern. The edges of the smoothed path (at this iteration) are shown in blue.}
    \label{fig:horseshoes}
\end{figure}

Each adjacent cell pair is horizontally, vertically or diagonally oriented, as shown in Figure~\ref{fig:smoothingcases}, and each orientation can result in the need to introduce additional adjacent cell pairs during smoothing. In the horizontal case, the consequence of the latitude correction is that the optimal crossing point selected between the entry and exit points can fall south of the bottom boundaries of one or both of the adjacent cells. When this happens, new cells must be added to the path. New cell pairs are always added in a {\em horseshoe pattern} consisting of as many as three adjacent cell pairs. Figure~\ref{fig:horseshoes} shows the possibilities that arise in the horizontal case, when both cells are the same size and when they are of different sizes as a result of cell-splitting. 

In the diagonal cases, the smoothed route will always need to pass through either the top or bottom cell on the missing diagonal, as shown by the dotted lines in Figure~\ref{fig:smoothingcases}c. In the vertical case, the crossing point must lie on the fixed latitude boundary between the cells, but if the entry and exit points are close to one of the longitudinal boundaries the crossing point can be pushed by currents beyond the boundary in the East or West direction. This results in the need for the addition of a horseshoe pattern.

\begin{figure}
    \centering
    \includegraphics[width=1\textwidth,keepaspectratio]{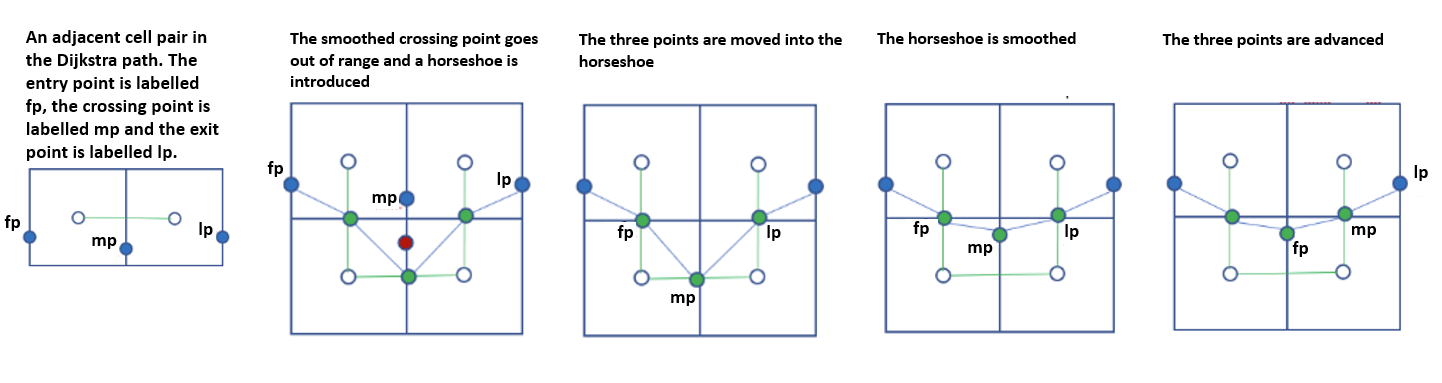}
    \caption{Smoothing a horizontal adjacent cell pair. The three tracking points (fp, mp and lp) are initialised in the leftmost pair, then updated as the horseshoe construction progresses.}
    \label{fig:horizontalcase}
\end{figure}

The smoothing process works by smoothing successive triples of points, called the first, mid and last {\em tracking points} ({\it fp, mp} and {\it lp}), and moving the tracking points along the path as smoothing progresses. Figure~\ref{fig:horizontalcase} shows the process by which an iteration works over a given horizontal adjacent cell pair in the Dijkstra path. The smoothing process is iterative, iterating over the entire path until there is convergence in the choice of crossing point in every adjacent cell pair visited. On each iteration, the whole path is considered again to see whether it can be smoothed further.

As the smoothing process nears convergence, no further pairs will be introduced but crossing points will continue to be moved until the shortest route (according to our curvature approximation) has been found.

\subsubsection{Horizontal Smoothing}
Using Equation~\ref{eqn:eqn5}, the speed of the vessel through the left hand cell is defined as: 
\begin{equation}
\label{eqn:eqn6nc}
s_l^2t_1^2 = (x - t_1u_1)^2 + (y - t_1v_1)^2
\end{equation}
 where $(u_1 + v_1)$ is the current vector $\vec{u_1}$ in the left cell, and $x$ and $y$ are the distances from the centre to the boundary and from the midpoint of the boundary to the $yval$, respectively. 
 
In the method described in section~\ref{sec:newtonsplitcells}, each cell is treated as having a constant width which is scaled by the latitude of its centre point. Now we want to introduce a latitude correction to remove this constant width assumption so that the choice of $yval$ more closely respects the curvature of the earth. We use a spherical geometry approximation, following \cite{equirect}~\footnote{\tt https://www.movable-type.co.uk/scripts/latlong.html}, in which we replace the cell widths, $x$ and $a$, with $x \,cos\, \theta$ and $a \cos \psi$, where $\theta = \frac{y}{R} + \lambda$ and $\psi = \frac{-b}{R}+ \tau$. Here, $\theta$ is the latitude of the $yval$, $\lambda$ is the latitude of the entry point into the left cell and $\tau$ is the latitude of the exit point on the right cell edge. Considering travel through the left cell, we have:
\begin{equation}
\label{eqn:new1}
s_l^2t_1^2 = (x \,cos\, \theta - t_1u_1)^2 + (y - t_1v_1)^2
\end{equation}
This approximation treats a cell as flat, so that the distance calculation in each cell can still be done using Pythagoras, with just one cosine operation rather than the multiple trigonometric functions required for calculating the haversine as would be required in a spherical geometry.

As in section~\ref{sec:newtonsplitcells}, we use the shorthand $\partial$ to mean $\frac{d}{dy}$. We use $z_l = x \,cos\, \theta$ to refer to the horizontal distance travelled in the left cell, and $z_r = a \,cos\, \psi$ to refer to the horizontal distance travelled in the right cell. This latitude correction is pushed through all of the equations~\ref{c1}-~\ref{eqn:t2diff}, so that:

\begin{equation}
d_1^2 = z_l^2 + y^2
\end{equation}
\begin{equation}
d_2^2 = z_r^2 + (Y-y)^2
\end{equation}

\begin{equation}
D_1 = z_lu_1 + yv_1
\end{equation}
\begin{equation}
D_2 = z_ru_2 + (Y-y)v_2
\end{equation}

$F(y)$, the function we wish to minimise, is:

\begin{equation}
\label{minf}
F(y) = (X_1 + X_2)y - t_1X_2v_1 + t_2X_1v_2 - YX_1 + \partial{z_r}(z_r - t_2u_2)X_1 + \partial{z_l}(z_l - t_1u_1)X_2
\end{equation}

The derivative of $F(y)$ is:

\begin{equation}
\label{solution}
\begin{array}{ll}
\partial{F(y)} = (X_1 + X_2) + y(\partial{X_1} + \partial{X_2})  \\
\quad \quad \quad- \frac{v_1}{C1}(\partial{X_2}(X_1 - D_1) + X_2(\partial{X_1}-\partial{D_1}))\\
\quad \quad \quad+ \frac{v_2}{C_2}(\partial{X_1}(X_2 - D_2) + X_1(\partial{X_2}-\partial{D_2})) - Y\partial{X_1}\\
\quad \quad \quad- \frac{z_r}{R^2}(z_r - \frac{(X_2 - D_2)u_2}{C_2})X_1 - \frac{z_l}{R^2}(z_l - \frac{(X_1 - D_1)u_1}{C_1})X_2\\
\quad \quad \quad+ \partial{z_r} (\partial{z_r}- \frac{u_2}{C_2}(\partial{X_2} - \partial{D_2}))X_1\\
\quad \quad \quad+\partial{z_l}(\partial{z_l} - \frac{u_1}{C_1}(\partial{X_1} - \partial{D_1}))X_2\\
\quad \quad \quad+\partial{z_r}(z_r - \frac{(X_2 - D_2)u_2}{C_2})\partial{X_1} +\partial{z_l}(z_l - \frac{(X_1 - D_1)u_1}{C_1})\partial{X_2}
\end{array}
\end{equation}

where $\partial{X_1}$ and $\partial{X_2}$ are defined as follows.

\begin{equation}
\label{auxs}
\begin{array}{ll}
\partial{X_1} = \frac{D_1\partial{D_1} + C_1(z_l\partial{z_l}+y)}{X_1}\\
\\
\quad\quad = \frac{(\partial{z_l}u_1+v_1)D_1 + C_1(y + z_l\partial{z_l})}{X_1}\\
\\
\quad\quad = \frac{D_1v_1+C_1y + \partial{z_l}(D_1u_1 + C_1 z_l)}{X_1}
\end{array}
\end{equation}

\begin{equation}
\begin{array}{ll}
\partial{X_2} = \frac{D_2(\partial{z_r}u_2 - v_2) + C_2(z_r\partial{z_r}-Y+y)}{X_2}\\
\\
\quad\quad = \frac{-v_2D_2 - C_2(Y-y) + \partial{z_r}(D_2u_2 + C_2z_r)}{X_2}\\
\end{array}
\end{equation}

Using equations~\ref{minf} and~\ref{solution}, we can use Newton's method to arrive at a value of $y$ that makes $F(y) = 0$. Due to the new latitude correction, this value of $y$ might drop further south than the latitudes of the bottom boundaries of the left and right cells. This results in the addition of a suitable horseshoe pattern, which will result in new cell pairs to be smoothed on later iterations.

\begin{figure}
    \centering
    \includegraphics[width=1\textwidth,keepaspectratio]{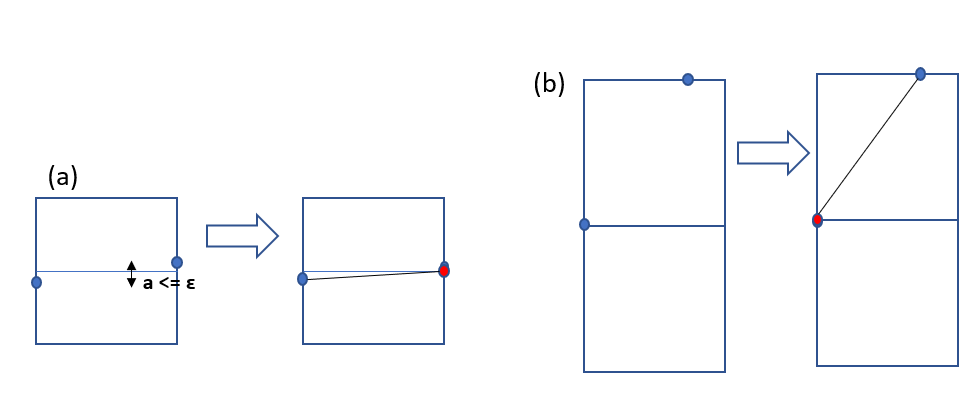}
    \caption{Smoothing special cases: (a) the entry and exit points are within $\epsilon$ of each other and in order to avoid convergence issues the $yval$ (red node) is placed on the longitudinal boundary closest to the exit point. (b) The entry point has moved to the boundary, so the left-hand travel time is zero. The $yval$ is placed over the entry point (by a symmetric argument, the exit point could move to the boundary so that the travel time in the right cell is zero and the $yval$ is placed over the exit point.)}
    \label{fig:smoothingspecialcase}
\end{figure}

During smoothing, it can happen that the travel time is zero in one or other cell. This can occur when, on some iteration during the smoothing process, the entry point has been moved onto the edge between the adjacent cells. If this happens, it is treated as a special case, in which the path simply tracks the edge of the cell. To aid convergence, this solution is used whenever the entry and exit points are within some small $\epsilon$ of each other, as shown in Figure~\ref{fig:smoothingspecialcase}.

\subsubsection{Vertical Smoothing} 

The derivation above deals with smoothing the crossing between two horizontally adjacent cells, in which the latitude of the crossing point varies. The journeys to and from the crossing point are determined by the latitudes of the entry and exit points. When travelling vertically, the latitude of the crossing point is fixed, and it is the longitude that varies. Also, travel to and from the crossing point is in the longitudinal direction so is not affected by the latitudes of the entry and exit points. Therefore the vertical case is different from the horizontal case and requires a different method for computing the travel time between the cells.

We now describe how the passage between vertically aligned cells is smoothed. 
When crossing from an entry point to an exit point in a vertical alignment, we will pass through a midpoint at a latitude, $\theta$, between the entry and exit point latitudes, $\lambda$ and $\psi$ respectively.

\begin{figure}
\centerline {\includegraphics[width=10cm]{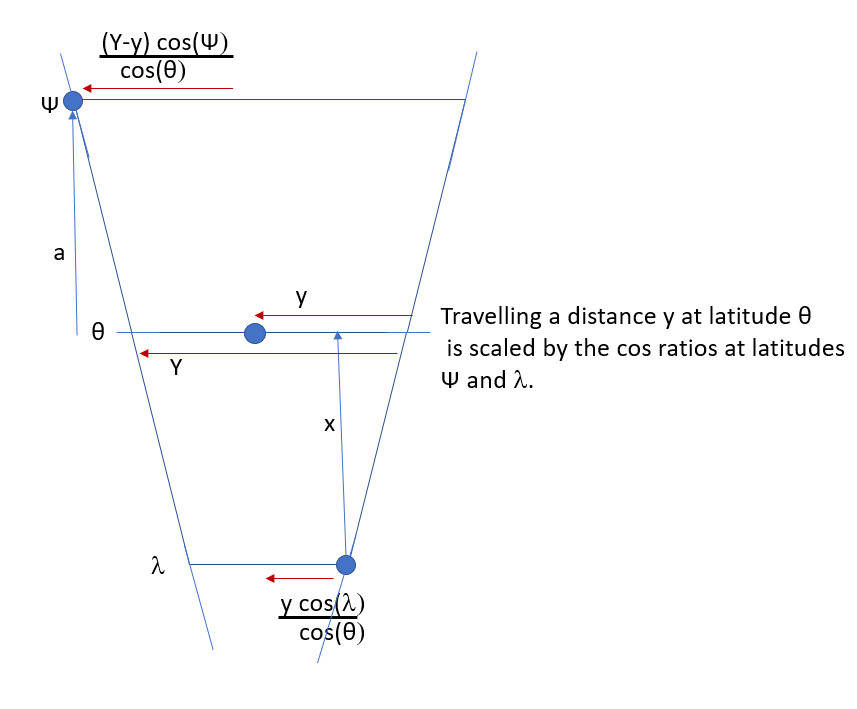}}
\caption{Using cos ratios to approximate horizontal distance in a vertical alignment.}
\label{cosrats}
\end{figure}

Suppose that the vertical distance from the entry point to the mid point is $x$ when travelling North, and $-x$ when travelling South, and that the vertical distance from the mid point to the exit point is $a$ ($-a$). When travelling from the side of a cell to its mid point at a latitude $\theta$, the distance travelled, $d$, depends on both the distance from the edge to the mid point and $\theta$. We approximate $d$ quite coarsely using cos ratios as shown in Figure~\ref{cosrats}. This shows two longitudinal lines and three latitudes, in a vertical adjacent cell pair arrangement. The entry point is at latitude $\lambda$, the exit point is at latitude $\psi$, and the boundary between the cells is at latitude $\theta$. As shown, a distance travelled at the boundary latitude must be scaled to give the corresponding distances at the other two latitudes. Two ratios are required, one to approximate the horizontal distance travelled at the entry latitude, and the other to approximate the horizontal distance travelled at the exit latitude. We call these ratios $r_1$ and $r_2$ respectively, defined in equations~\ref{leftcosratio} and~\ref{rightcosratio}.

\begin{equation}
\label{leftcosratio}
r_1 = \frac{cos \lambda}{cos \theta}
\end{equation}

\begin{equation}
\label{rightcosratio}
r_2 = \frac{cos \psi}{cos \theta}
\end{equation}

\begin{equation}
\label{littled1r}
d_1^2 = x^2 + (r_1y)^2
\end{equation}

\begin{equation}
\label{littled2r}
d_2^2 = a^2 + (r_2(Y-y))^2
\end{equation}

\begin{equation}
\label{bigd1r}
D_1 = xu_1 + r_1v_1y
\end{equation}

\begin{equation}
\label{bigd2r}
D_2 = au_2 + r_2v_2(Y-y)
\end{equation}

\begin{equation}
\partial{D_1} = r_1v_1
\end{equation}

\begin{equation}
\partial{D_2} = -r_2v_2
\end{equation}

The travel times in the start and end cells, $t_1$  and $t_2$, are given as: 

\begin{equation}
\label{tt1}
s_l^2t_1^2 = (x - t_1u_1)^2 + (r_1y - t_1v_1)^2
\end{equation}

\begin{equation}
\label{tt2}
s_r^2t_2^2 = (x - t_2u_2)^2 + (r_2(Y-y) - t_2v_2)^2
\end{equation}

Differentiating equations~\ref{tt1} and~\ref{tt2}, we obtain:
\begin{equation}
s_l^2t_1\partial{t_1} = -(x - t_1u_1)u_1\partial{t_1} + (r_1y - t_1v_1)(r_1 - v_1\partial{t_1})
\end{equation}

\begin{equation}
\partial{t_1} (s_l^2t_1 + (x - t_1u_1)u_1 + v_1(r_1y - t_1v_1)) = r_1(r_1y - t_1v_1)
\end{equation}

\begin{equation}
\partial{t_1} ((s_l^2 - u_1^2 - v_1^2)t_1 + x u_1 + r_1v_1y) = r_1(r_1y - t_1v_1)
\end{equation}

\begin{equation}
\partial{t_1} (C_1t_1 + D_1) =  r_1(r_1y - t_1v_1)
\end{equation}

\begin{equation}
\partial{t_1} X_1=  r_1(r_1y - t_1v_1)
\end{equation}

\begin{equation}
\partial{t_1} = \frac{ r_1(r_1y - t_1v_1)}{X_1}
\end{equation}

and, by similar reasoning:

\begin{equation}
\partial{t_2} = \frac{ r_2(r_2(Y-y) - t_2v_2)}{X_2}
\end{equation}

As before, we require $\partial{t_1} + \partial{t_2} = 0$.

We differentiate $X_1$ and $X_2$ to obtain:

\begin{equation}
\partial{X_1} = \frac{r_1 (D_1v_1 + r_1C_1y)}{X_1}
\end{equation}

and 
\begin{equation}
\partial{X_2} = \frac{-r_2 (D_2v_2 + r_2C_2(Y-y))}{X_2}
\end{equation}

We can now write:
\begin{equation}
\partial{t_1} + \partial{t_2} = r_1(\frac{r_1y - t_1v_1}{X_1}) + r_2(\frac{r_2(y - Y)+t_2v_2}{X_2}) = 0
\end{equation}

This can be rewritten as follows using equations~\ref{t1a} and~\ref{t2a}:
\begin{equation}
\label{fy}
F(y) = (r_2^2X_1 + r_1^2X_2)y - \frac{r_1(X_1 - D_1)X_2 v_1}{C_1} + \frac{r_2(X_2 - D_2)X_1v_2}{C_2} - r_2^2YX_1
\end{equation}

which, differentiated, gives:

\begin{equation}
\label{deltafy}
\begin{array}{ll}
\partial{F(y)} = (r_2^2X_1 + r_1^2X_2) + y(r_2^2\partial{X_1} + r_1^2\partial{X_2})\\
\quad \quad  - \frac{r_1v_1}{C_1}((\partial{X_1} - \partial{D_1})X_2 + (X_1 - D_1)\partial{X_2})\\
\quad \quad + \frac{r_2v_2}{C_2}((\partial{X_2} - \partial{D_2})X_1 + \partial{X_1}(X_2 - D_2))\\
\quad \quad - r_2^2Y\partial{X_1}
\end{array}
\end{equation}

Using the definitions of $F(y)$ and $\partial{F(y)}$, equations~\ref{fy} and~\ref{deltafy}, we use Newton's method to arrive at a value of $y$ that makes $F(y) = 0$. It can happen that the chosen $y$ falls outside one of the longitudinal boundaries due to the strength of the current present in the cell. This will result in the addition of a horseshoe pattern to enable smooth passage between the entry and exit points. Futhermore, as in the horizontal smoothing case, it can happen that an iteration of smoothing moves the entry point to within $\epsilon$ of the boundary between the cells. In this case the special case solution shown in Figure~\ref{fig:smoothingspecialcase} is used.

\subsubsection{Dealing with Special Cases in Smoothing}
\label{smoothingspecialcases}
If completely unconstrained, the smoothing process can lead the route far away from the optimal Dijkstra path. If this happens, the resulting smoothed route might visit cells that are actually inaccessible because of ice or land, or that represent much worse conditions than the cells visited by the Dijkstra path. This could happen because the smoothing process only has a local view of the adjacent cell pair under consideration (and any neighbouring cells that must be added when the crossing point falls out of range) and does not have any global optimisation strategy. We want to avoid this, because the Dijkstra path represents the optimal route in the mesh. Smoothing should remove the dependency on the mesh artifacts, thereby typically shortening the path and making it more fuel-efficient, whilst adhering to the guidelines of the Dijkstra path.

We introduce three methods for constraining the smoothing process.

\begin{figure}[t!]
    \centering
    \includegraphics[width=0.8\textwidth,keepaspectratio]{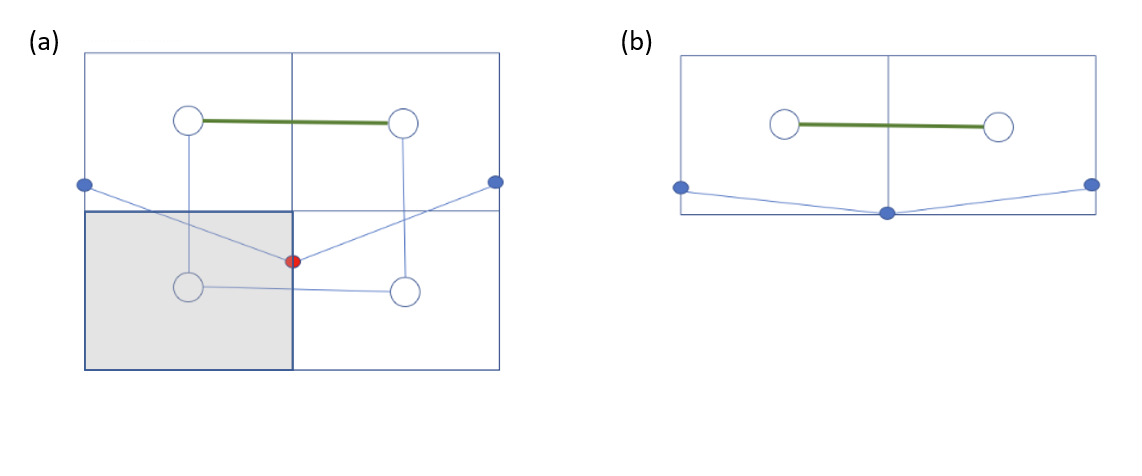}
    \caption{(a) One or both of the two new cells are blocked by ice or land (the grey cell). (b) The crossing point is clipped back to the original cell pair.}
    \label{fig:clip}
\end{figure}

\begin{figure}[t!]
    \centering
    \includegraphics[width=1\textwidth,keepaspectratio]{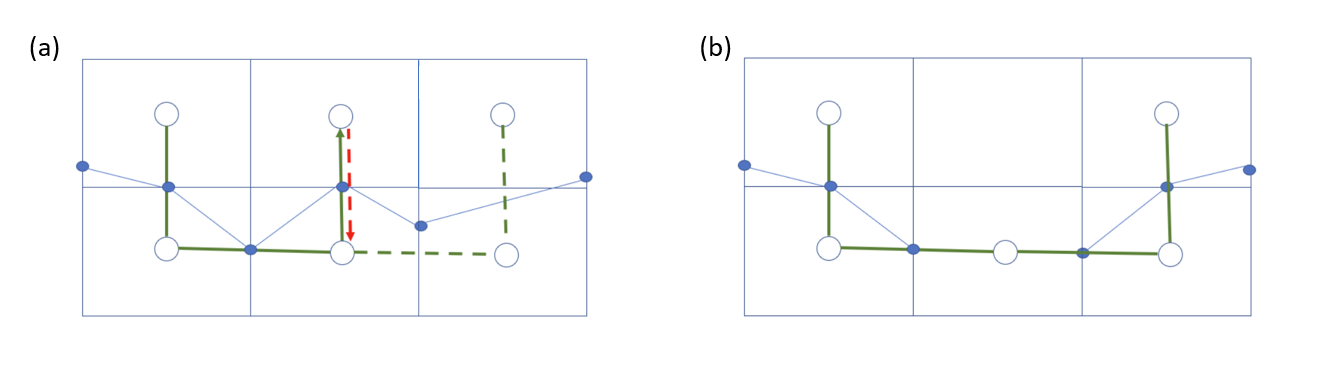}
    \caption{(a) Introduction of a second horseshoe can produce reversed edges (the red dotted line and the green line in the middle cell reverse one another). (b) These are removed.}
    \label{fig:revedges}
\end{figure}

\begin{enumerate}
\item When constructing a horseshoe, as shown in Figure~\ref{fig:horseshoes}, the crossing point is clipped back to a corner of the original adjacent cell pair if either of the newly accessible cells are blocked by extreme ice concentration or land (for example, see Figure~\ref{fig:clip}).
\item Whenever new adjacent cell pairs are introduced, any reversing edges that have been introduced are removed (for example, see Figure~\ref{fig:revedges}). This prevents the smoothing process from introducing unnecessary loops in the path.
\item We do not allow smoothing to enter cells that have a much higher ice concentration than the cells in the original adjacent cell pair. Entering these cells would introduce step changes in the vessel speed, power requirements and fuel demand which would be infeasible to execute in reality. Instead we clip the crossing point back to the original cell box pair, as in (1) above.
\end{enumerate}

\subsubsection{Convergence of the Smoothing Process}
\begin{figure}[t!]
    \centering
    \includegraphics[width=1.0\textwidth,keepaspectratio]{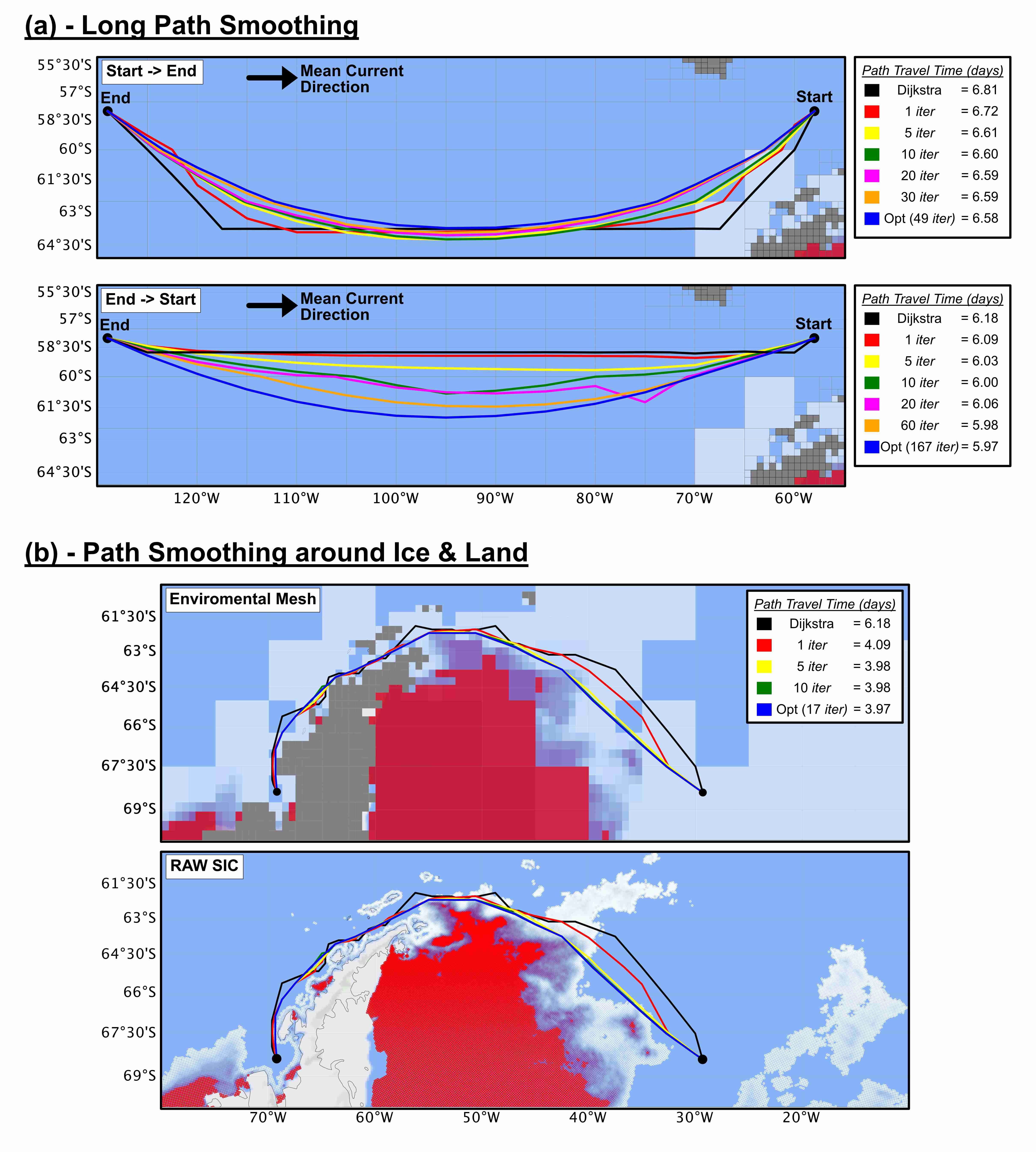}
    \caption{Example Path Smoothing over great distance and around Ice/Land. (a) path smoothing over great distances. (b) smoothing around land and ice.} 
    \label{fig:SmoothingLongpath}
\end{figure}

Smoothing tends to converge within at most a few hundred iterations. We set an upper bound of 1000 iterations. Convergence occurs when two successive paths are less than $10^{-3}$days different in travel time. The convergence of the smoothing process for two examples can be seen in Figure \ref{fig:SmoothingLongpath}. 

\section{Results}\label{sec-results}
In this section we apply our methods to the British Antarctic Survey (BAS) research and supply vessel, the RRS Sir David Attenborough (SDA), and validate the planned routes generated in this case.

\begin{figure}[h]
    \centering
    \includegraphics[width=1.0\textwidth,keepaspectratio]{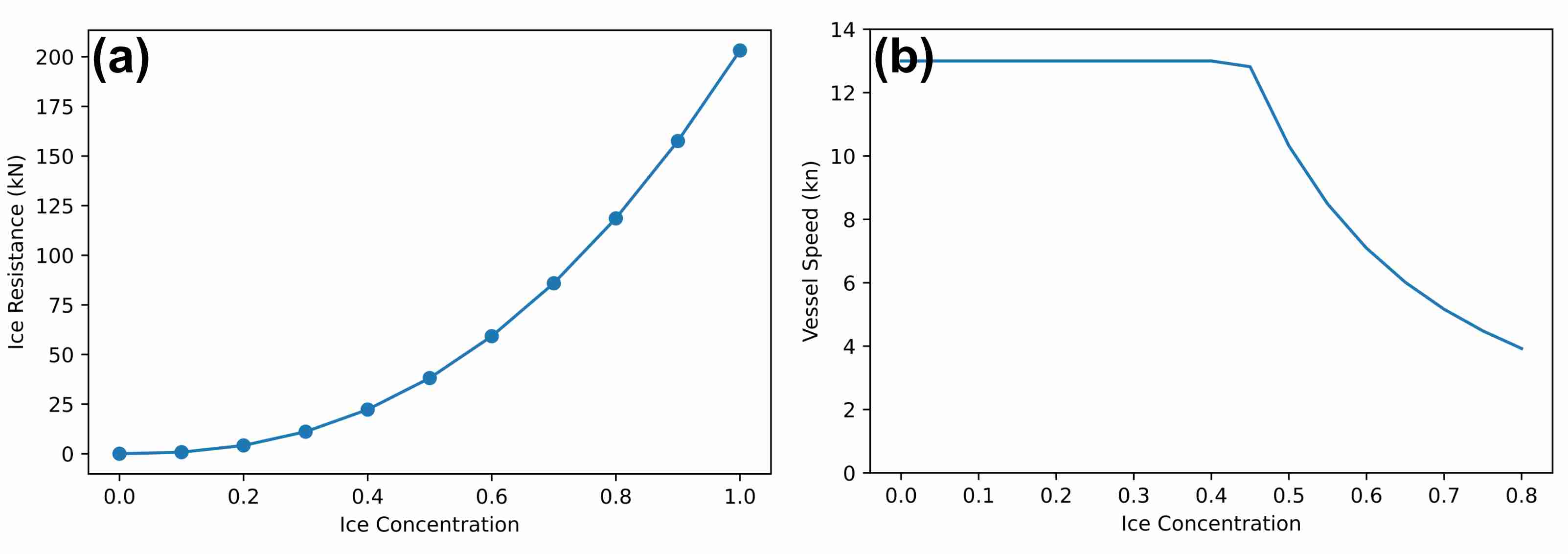}
    \caption{(a) The ice resistance force acting on an ice breaking vessel for different ice concentration values. (b) the recommended speed for a given ice concentration.} 
    \label{fig:iceshipinfo}
\end{figure}

The SDA is a Polar Class 5-rated ice-strengthened vessel and is therefore designed to be capable of breaking medium thickness first year ice (approximately 1m thickness). The SDA underwent its maiden voyage in the Southern Ocean in January 2022. Relatively little observational data is so far available, so our modelling of ice resistance and fuel requirements are currently based on summary figures rather than on measured experience. We have applied the modelling technique described in Section~\ref{sec-fuel} to a set of summary fuel consumption figures for different operating modes of the SDA. For the following examples we have chosen a conservative maximum average ice concentration threshold of 80\%, meaning that a cell with an average ice concentration of more than 80\% will not be entered. The maximum speed achievable by the SDA is taken to be its open water cruising speed of 13 knots (1 knot = 1.852~km/h). Given these parameters we can now calculate the ice resistance force as a function of the ice concentration. The result of this can be seen in Figure~\ref{fig:iceshipinfo}a.

An example of the speed variation algorithm in use is shown in Figure~\ref{fig:iceshipinfo}b. This shows the speed recommended for a vessel with similar parameters to the SDA at different ice concentrations when travelling through level ice with a thickness of 0.8~m. The force limit used to determine when to slow the vessel is derived from the ice breaking specification of the SDA which states that it can achieve a speed of 3 knots when breaking 1m of level ice.

In order to determine the fuel usage, at any speed and in any ice conditions, a polynomial fit was made to these summary figures. The resulting equation for daily fuel use is given by Equation~\ref{eq:fuelmodel}. This model rests on the assumption that the ship consumes a certain amount of fuel to achieve a given speed, determined from the open water figures, and that it requires an additional quantity of fuel in order to overcome the ice resistance force and maintain that speed, determined from the ice breaking figures. There is also a base fuel consumption of 6 tons per day for essential services onboard (heating, lighting etc). Equation~\ref{eq:fuelmodel} gives the vessel's fuel usage, $U_d$, in tons per day (t/day) for a given speed in knots (kn) and a given ice resistance force in kilo Newtons (kN). \\

\begin{equation}
    U_{d} = 0.113V^2 - 0.132V + 0.003R^{2} + 0.042R + 6.0 
\label{eq:fuelmodel}
\end{equation}

This model is simple, and will be further developed in future work as the SDA gains experience in operating in ice, and more data becomes available.

\subsection{Path Validation}
In this section we investigate and evaluate the paths generated for the SDA vehicle for varying sea-ice concentrations, exploring the stability of the path generation results under changing environmental and user-defined conditions. 

Throughout this section all routes are based on a mesh of the AMSR-2 Sea Ice Concentration spanning 2019-2020, unless stated otherwise. A temporal abstraction of 14 days is used, with a maximum splitting level of 3. 

    

\subsubsection{Quality of the Smoothed Routes}
The Dijkstra path construction inherently determines the optimal route through the meshed environment. However, smoothing these paths off the mesh relaxes the formulation and sacrifices the optimality guarantee, requiring evaluation of the quality of the resulting paths. We expect smoothing to reduce the travel time implied by a path, and in fact, in open water, the smoothing process as mathematically defined will monotonically improve the routes. However, we smooth around and into ice cells as well, which involves applying ice tolerance thresholds in a heuristic way. For example, as discussed in section~\ref{smoothingspecialcases}, when building horseshoes we disallow the addition of cells that have high enough ice concentrations to slow the vehicle down significantly or demand significant extra fuel by comparison with the original Dijkstra path. We therefore need to determine whether smoothed paths truly represent a reduction in travel-time over the meshed Dijkstra path. 

\begin{figure}
    \centering
    \includegraphics[width=1.0\textwidth,keepaspectratio]{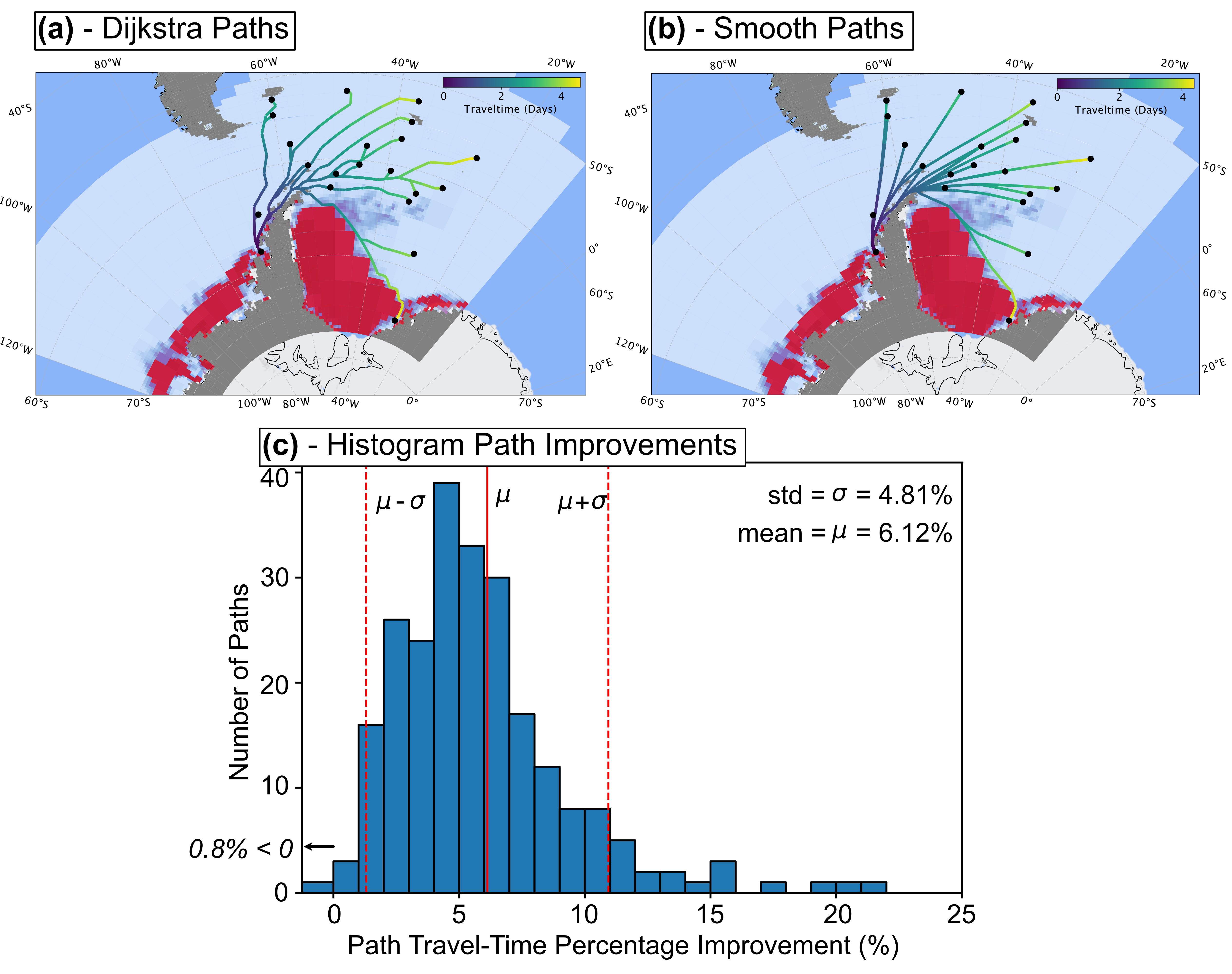}
    \caption{Histogram showing the travel-time percentage improvement of the smoothed path construction over the original Dijkstra paths.}
    \label{fig:results_pathhistogram}
\end{figure}

In order to evaluate this we construct Dijkstra paths and subsequent smoothed paths between a series of waypoints (Supplementary Table \ref{tab:waypoints}) across a range of seasonal sea-ice concentrations. An example of a subset of all the constructed Dijkstra and smoothed paths from Marguerite Bay to all other waypoints are shown in Figure \ref{fig:results_pathhistogram}a and b respectively. Once all paths are constructed the total travel-time of the Dijkstra and smoothed paths is determined for all waypoint pairs, with the travel-time percentage improvement from Dijkstra to smoothed path determined. The histogram of the path improvements is given in Figure \ref{fig:results_pathhistogram}, demonstrating a mean improvement across all paths of $6.12\%$ with a standard deviation of $4.81$ hours and only $0.8\%$ of the paths show no improvement. 

\subsubsection{Time series averaging}

To test the robustness of the generated routes against the real conditions, we examine the effect of time-averaging the ice-concentration for route-planning by running the generated paths against the original unprocessed data. In order to quantify how well the routes perform when run against the original data we determine a {\em path violation} to have occurred whenever the path comes close to an ice point that exceeds the concentration that is navigable by the vehicle (e.g. $> 80\%$ for the SDA example). Based on this evaluation, we can select the best abstraction to use to balance the computational cost of route construction with the robustness of the resulting routes.

Route-planning with a temporal abstraction of $k$ days means that conditions over the whole route are assumed to be described by the $k$-day average of the abstraction. If $k$ is much smaller than the length of the route, then route-planning is blind to changes in the true conditions and path violations will occur when a route is executed. If $k$ is much larger than the length of the route, averaging will lose too much information and path violations will occur. We perform a number of experiments to evaluate the extent of path violations under different abstractions. 

 To initialise the testing procedure we first resample the routes at $10$km spacing, so as not to bias paths transiting shorter distances. Once re-sampled, a 15$km$ {\em radius of influence} is determined around each path point, determining the mean and standard-deviation of the original unprocessed ice concentration data for the date on which that point is visited on the path. This allows us to determine if any point along the path constitutes a violation. A series of increasingly coarse temporal abstractions is selected for the construction of the mesh, represented by abstracting 2, 7, 14, 28 and 56 days. Routes are then planned in these meshes for all waypoint pairs for every month of 2019, with the temporal abstractions starting on the first day of each month.

\begin{figure}
    \centering
    \includegraphics[width=0.8\textwidth,keepaspectratio]{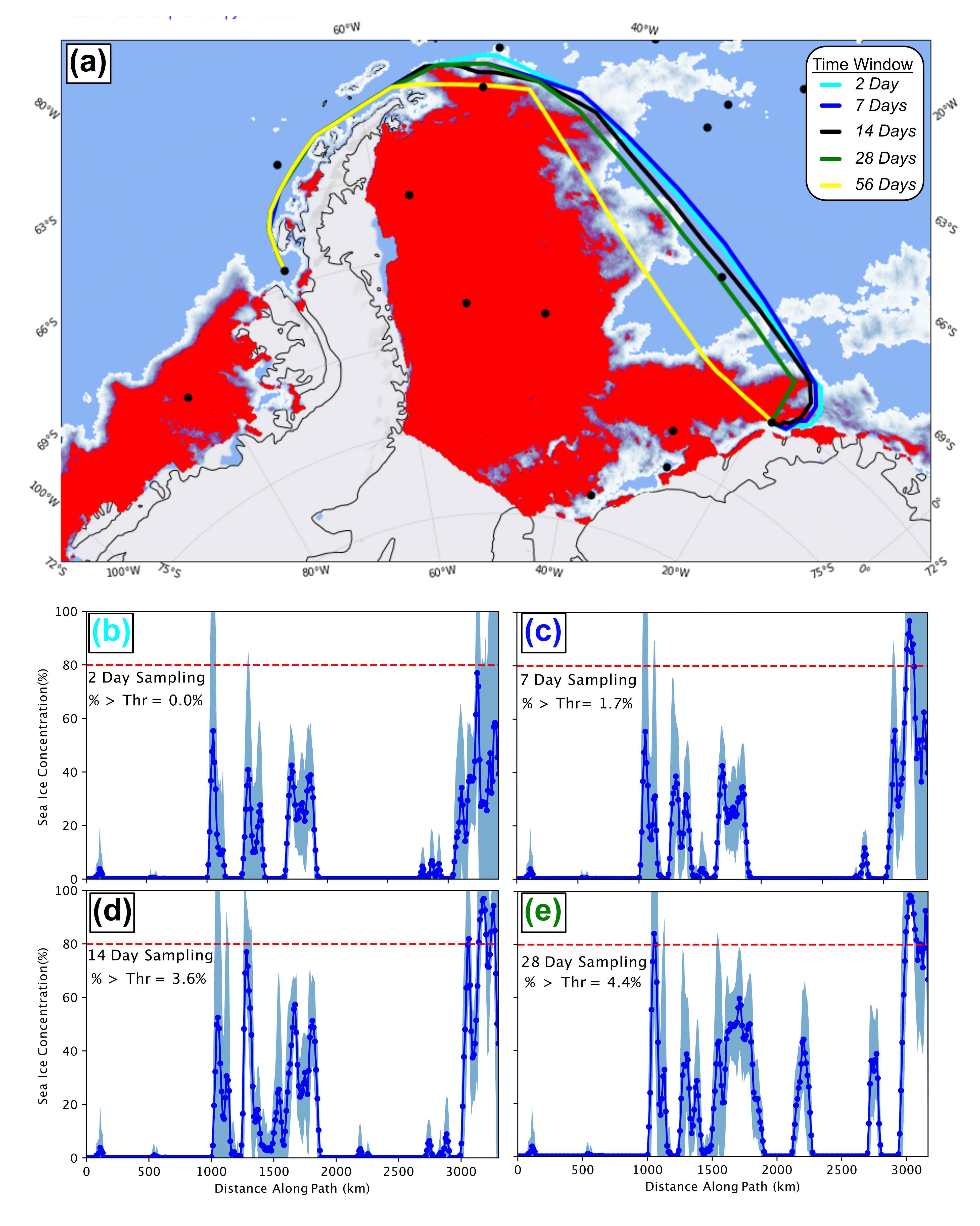}
    \caption{Time series averaging effects on the smoothed route between Marguerite Bay and Sassi Moorings with sea-ice concentration starting 1'st January 2020. (a) represents the smooth routes for different abstraction sizes. The purple colourmap represents the unprocessed 2-day average of the AMSR-2 SIC, with the red-regions representing the SIC above that navigable by the vessel. Black points represent user defined waypoints. (b) - (e) represent the cross-sections along the routes with label colour corresponding to abstraction size. The dark-blue dotted line represents the unprocessed mean SIC within the radius of influence and blue region the 2-std region of the SIC. Red dashed line represents the maximum extent SIC for the vessel.}
    \label{fig:results_timevaryingpaths}
\end{figure}

 For example Figure \ref{fig:results_timevaryingpaths} demonstrates the single waypoint pair from Marguerite Bay, a waypoint situated close to the BAS Research Station
Rothera, to SASSI Moorings, a location of scientific interest on the East coast of the Weddell Sea. The time windows corresponding to the temporal abstractions all start on 1st January 2020. Panel a represents the paths generated under the different temporal abstractions, overlaid on the 2-day averaged unprocessed sea-ice concentration. In this example a series of the corresponding cross-sections is given in Figures \ref{fig:results_timevaryingpaths}b-e, with each cross-section showing the mean (blue line) and 2-std confidence (blue region) of the unprocessed ice concentration data within the radius of influence. These cross-sections demonstrate that increasing the coarseness of the temporal abstraction increases the percentage of points along the routes in violation, for example the 7 day abstraction shows that 3.6\% of the path points are in violation, while this rises to $4.4\%$ under the 28-day abstraction. 
 
\begin{figure}
    \centering
    \includegraphics[width=1.0\textwidth,keepaspectratio]{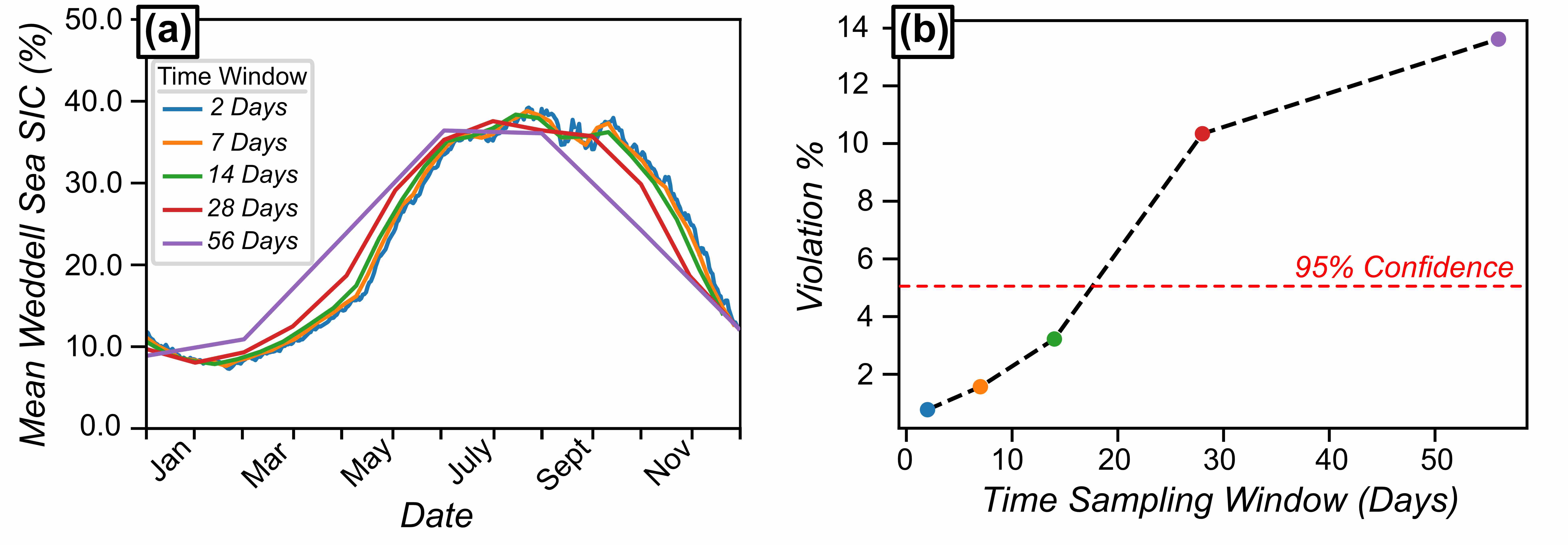}
    \caption{Temporal abstraction size for the mesh creation and its effect on Mean SIC and \% violation of path construction relative to the unprocessed SIC data. (a) represents the mean Weddell SIC seasonal variations and the effect of the temporal abstraction size on the mesh construction. (b) represents the percentage of path points across all routes in the whole year that are in violation. The colour of the points corresponds to the temporal abstraction used, with temporal abstractions finer than 18 days found to be within the 95\% confidence interval.}
    \label{fig:results_timeValidation_CornerPlot}
\end{figure}

 Applying this to all waypoint pairs we can determine the global percentage violation of path points across all paths relative to the temporal abstraction (Figure \ref{fig:results_timeValidation_CornerPlot}b), determining that a temporal abstraction of $18$ days or less is within the $95\%$ confidence interval. As most of our routes are less than 14 days in length, we choose a $14$-day sampling period as the optimal consistent with the $14$-day SIC mesh window shown in Figure~\ref{fig:results_timeValidation_CornerPlot}a. Using this abstraction is equivalent to assuming that conditions never change during any route. Of course, conditions do change. One way to address this is by dynamically re-planning when changes are observed or forecast during navigation of a planned route, and rebuilding both the mesh and route during a dynamically initiated re-planning phase.

\subsection{Example Routes}
In this section we demonstrate the use of the route planner in the construction of routes from different starting waypoints. This section is split into three parts. We first show the results of route-planning in high- and low-SIC years, in order to show the seasonal variation in the constructed routes. We then show how routes can be routes can be optimised according to different objective functions. Finally, we show the versatility of the route planner by constructing routes in different geographic locations.

\subsubsection{Seasonal Variations in Routes}
Sea ice extent is an ever-changing condition in the Central Weddell Sea with 2019 showing the lowest SIC in the area (\cite{Jena2022}).
The changing conditions mean that very different routes are taken between waypoint pairs at different times of year, and in some cases of extreme SICs, typically inaccessible scientific waypoints can become reachable. Unlike previous route-planning methods, such as those cited in Section~\ref{sec:background}, the work described in this article can efficiently compute smoothed routes with changing environmental SICs allowing us to investigate the different routes for different legacy SICs as well as for the current or a projected year.

\begin{figure}
    \centering
    \includegraphics[width=1.0\textwidth,keepaspectratio]{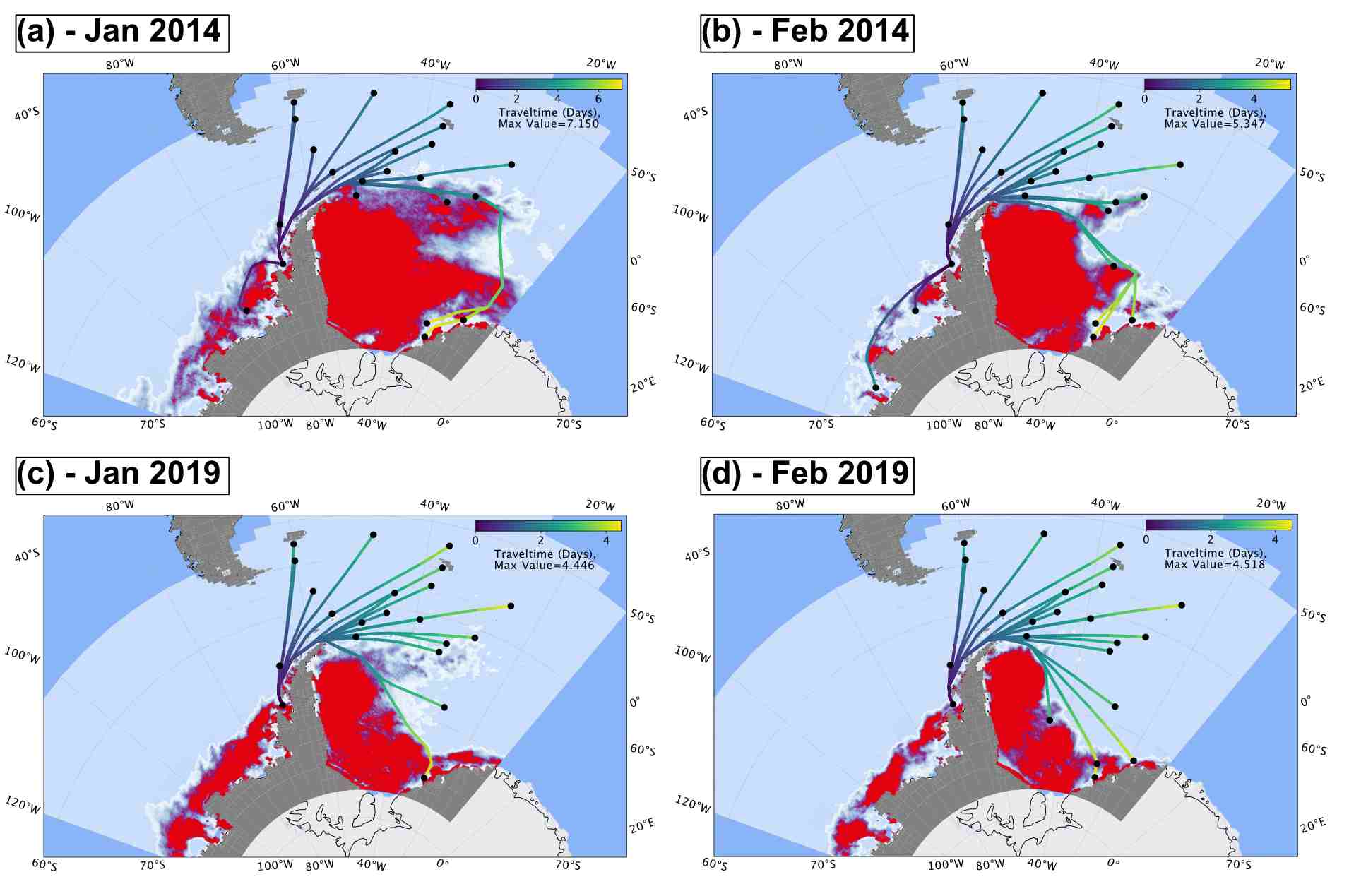}
    \caption{2014 and 2019 optimal travel-time route variation for paths from Marguerite Bay to all other waypoints during two operational months. Panels are separated into January and February, which represent the two main field season months in Antarctica. Black dots represent user defined waypoints. Coloured lines represent the optimal travel-time routes, where colour represents the travel-time in unit days. Colourmap represents SIC where darker purple is higher concentration, areas of red represent regions of SIC above that navigable by the vessel.}
    \label{fig:results_SeasonalPathVariation}
\end{figure}

One such example outlined in Figure \ref{fig:results_SeasonalPathVariation} shows the routes generated for the two main operational months (January and February) of 2014 and 2019. 2014 represents a higher than average SIC and 2019 represents the lowest mean SIC on record. It can be seen that in January 2014 time-optimal routes travelling in the Weddell Sea require a detour much further east compared to January 2019, requiring a 3-day increase in travel-time for the longer paths. These additional costs may outweigh the scientific benefits gained, which might justify deferring some scientific goals to a later year. Being able to explore how the fuel requirements and travel times of routes will vary under different sea ice conditions, and rapidly conduct comparative evaluations of routes, is an essential part of planning science cruises across a 2 or 3-year horizon.

\subsubsection{Fuel and Travel-time Optimised Routes}
\begin{figure}
    \centering
    \includegraphics[width=1.0\textwidth,keepaspectratio]{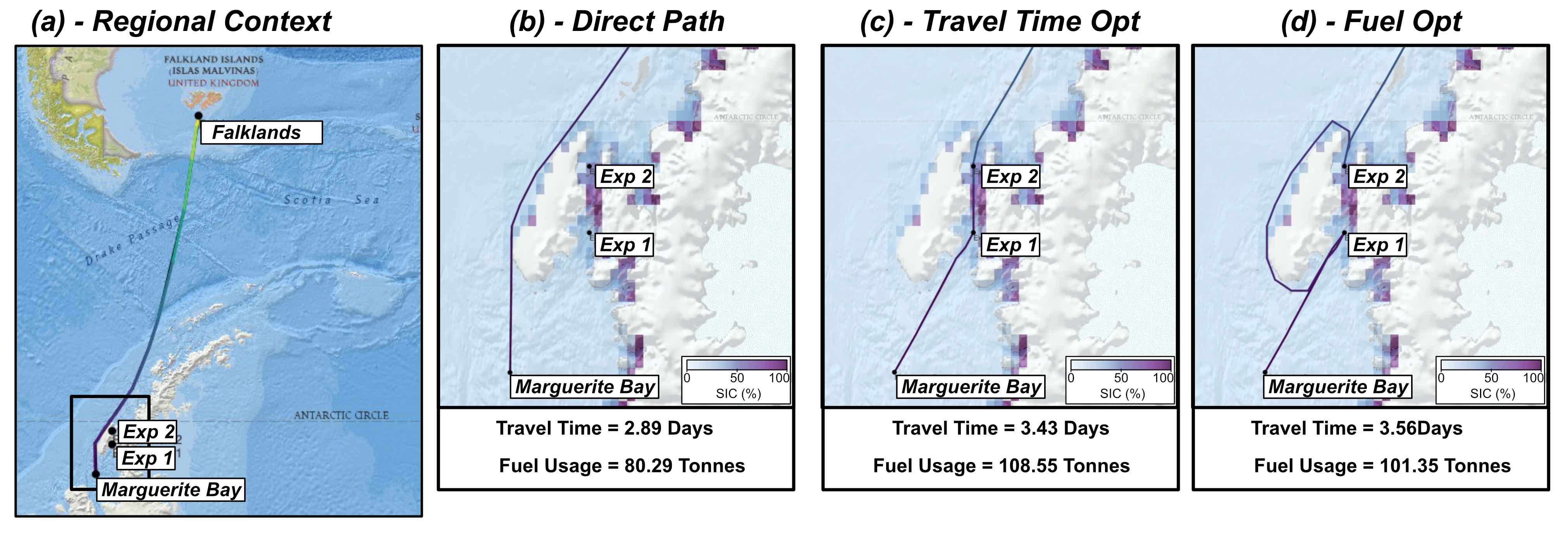}
    \caption{Example route planning with changing objective functions for the completion of two possible experiments (Exp 1 and Exp 2). (a) represents the travel time optimised route between Marguerite Bay and the Falklands for regional context. (b) represents a zoom in of panel a with SIC overlaid. (c) represents the travel time optimised route now undertaking Exp 1 and Exp 2 on the way. (d) represents the fuel optimised route now taking Exp 1 and Exp 2 underway.}
    \label{fig:FuelTravelTime}
\end{figure}

In Section \ref{sec:routeplanning} we observed that different user-defined objective functions can be used in route construction. In order to understand how different objective functions can affect the generated routes, we construct a simple example in which there are two waypoints, Marguerite Bay and the Falklands (Figure \ref{fig:FuelTravelTime}), with a requirement to plan a route between them. In addition, there are two locations at which science experiments can optionally be completed on the way. The locations are named `Exp1' and `Exp2'. The direct route optimised for shortest travel time from Marguerite Bay to the Falklands, without doing the experiments, would cost 2.88 days of travel and 80.29 tonnes of fuel; this acts as the benchmark of the minimum that can be achieved without doing the experiments. If we wish to do both of the experiments but still optimise the route for travel time then the quickest route would involve ice-breaking through the narrow channel known as the Gullet (demonstrated by the route going through the purple SIC in Figure \ref{fig:FuelTravelTime}c). This route would cost an additional 0.54 days of travel (12.96 hrs) and an additional 28.26 tonnes of fuel. It can be seen that the additional fuel expense for conducting the two experiments is considerable even though the additional cost in travel time is small. To explore the trade-off between travel time and fuel use, we construct the route that minimises fuel usage instead of time. The resulting route tracks around Adelaide island instead of going through the Gullet (Figure \ref{fig:FuelTravelTime}d). This fuel optimised route incurs an additional 0.13 days (3.12 hrs) of transit but requires 7.2 tonnes less fuel than the travel time-optimised route.

\subsubsection{Route Planning Beyond the Southern Ocean}
\begin{figure}
    \centering
    \includegraphics[width=1.0\textwidth,keepaspectratio]{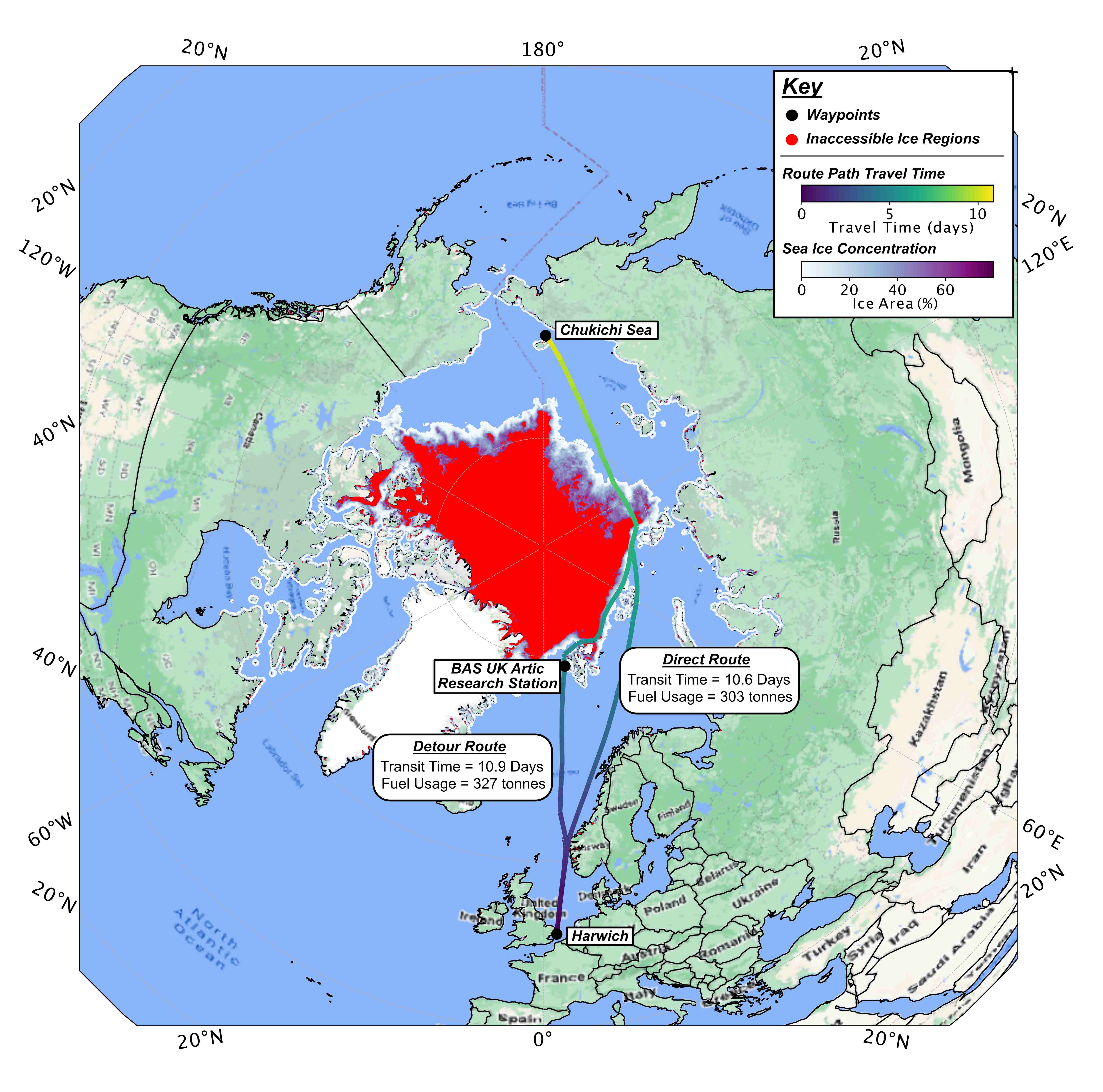}
    \caption{Arctic route planning example for Sept-2019. The colourlines represent the two travel-time optimised routes for the vessel travelling from Harwich to Chukchi Sea, with one route being the direct route and the other a detour via the NERC UK Arctic Research Station. The purple colourmap represents SIC, with red regions those inaccessible by the SDA.  This detour requires an additional 24 tonnes of fuel and a 0.4 day (7.2hr) increase in transit time.}
    \label{fig:articRoutePlanning}
\end{figure}

The route planning method described in this article has been demonstrated on navigation problems within the South Atlantic sector of the Southern Ocean, including the Weddell Sea region. However, the method is not limited to a specific region and depends only on the datasets used in the mesh construction. In order to demonstrate this generality we outline a test case for navigation within the Arctic Ocean, and we give an example for a navigation between Harwich (UK) and the Chukchi Sea, close to the Bering Strait. In this example we wish to understand the optimal travel-time route between these two waypoints, and find the additional costs in travel-time and fuel usage when the route between these two waypoints is required to visit the NERC UK Arctic Research Station, Ny-\AA lesund (Figure \ref{fig:articRoutePlanning}), on the way. To run this example requires only alterations to the input data files (we require the northern hemisphere AMSR-2 SIC, ocean currents and  ocean depths) and a new list of user defined waypoints. In this example we demonstrate that a detour via the NERC UK Arctic Research Station requires an additional 11 tonnes of fuel and 7 hours extra travel time.

We conducted experiments to see how our routes compare with those generated by~\cite{Mishra2021}, \cite{Kotovirta2009} and \cite{Lehtola2019}. These are the three approaches in the literature that are most similar to ours, as they are all based on building an informative mesh of the environment, modelling ice resistance information and using Dijkstra's algorithm to generate lowest cost paths. The work by~\cite{Lehtola2019} is particularly relevant because, as we do, they separate the route construction process into two separate stages: Dijkstra path construction and subsequent path refinement to improve geodesic validity. 

\begin{figure}
    \centering
    \includegraphics[width=0.9\textwidth,keepaspectratio]{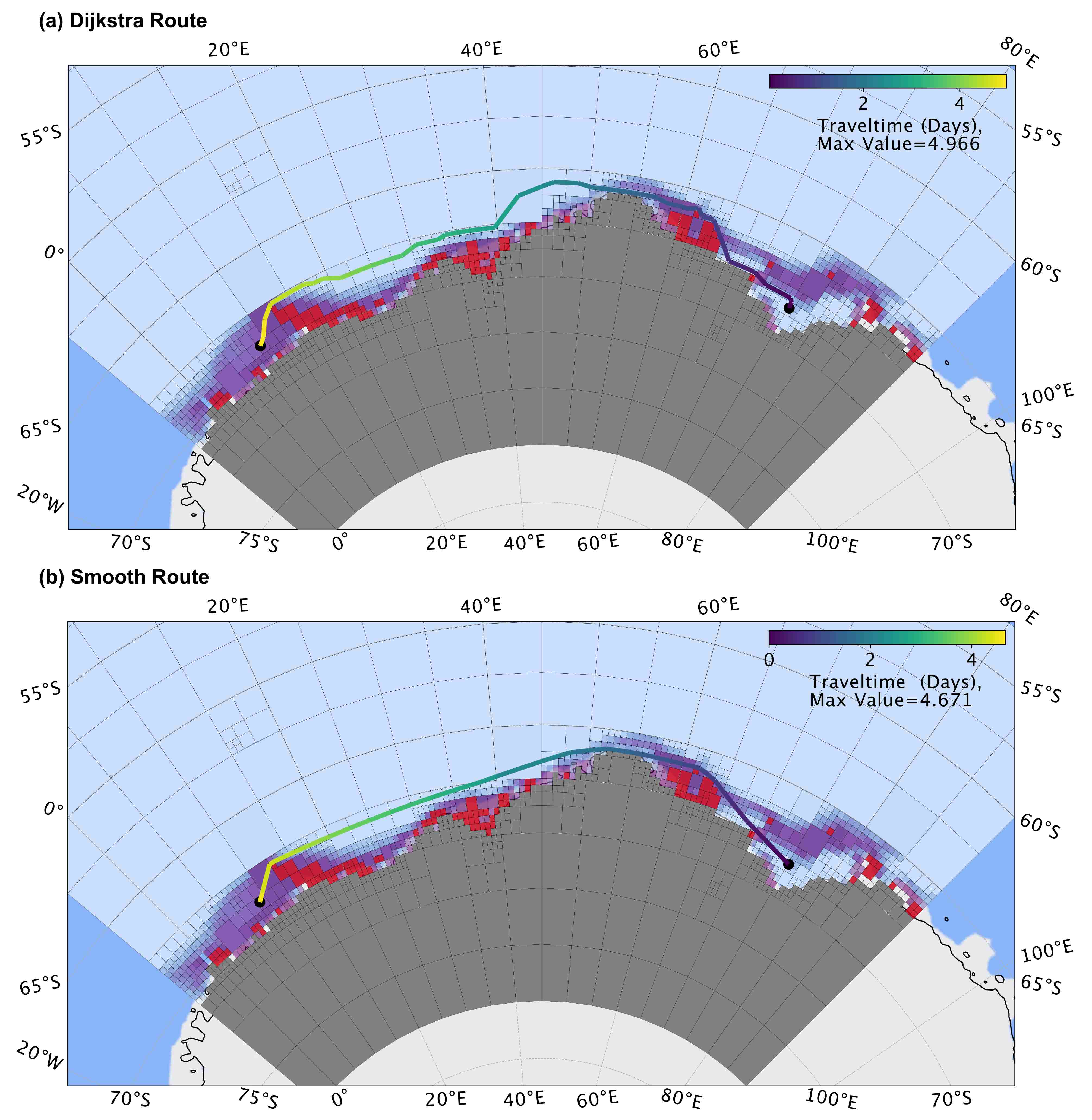}
    \caption{Comparison between our approach and that of \cite{Mishra2021} on the Bharati to Maitri route. (a) the travel-time optimised Dijkstra path that we generate in February 2019 ice. (b) the smoothed route in the same month. It can be seen that the smoothed path achieves a travel time of 4.671 days, an improvement of more than half a day over the path proposed by~\cite{Mishra2021}}.
    \label{fig:bharatimaitri}
\end{figure}

In Figure~\ref{fig:bharatimaitri} we show the paths that we construct, in February 2019, between Bharati and the Maitri (Indian) research station on the East coast of Antarctica. These paths can be compared with the optimised path shown in Figure 12 of~\cite{Mishra2021}. The comparison is not exact, because \cite{Mishra2021} model the effects of wind as well as ice, whilst our approach only models the SIC and surface ice extent. Wind off the coast could account for at least some of the discrepancy in  path length (our smoothed path is more than a day shorter than theirs). Despite these differences in modelling, it is clear that our routes gain a considerable advantage from the use of a non-uniform mesh coupled with smoothing. Despite our vessel being slower than theirs, our Dijkstra path, shown in  Figure~\ref{fig:bharatimaitri}a, is almost half a day shorter than theirs, and our smoothed path (part b) is 0.789 days shorter.

\begin{figure}
    \centering
    \includegraphics[width=0.8\textwidth,keepaspectratio]{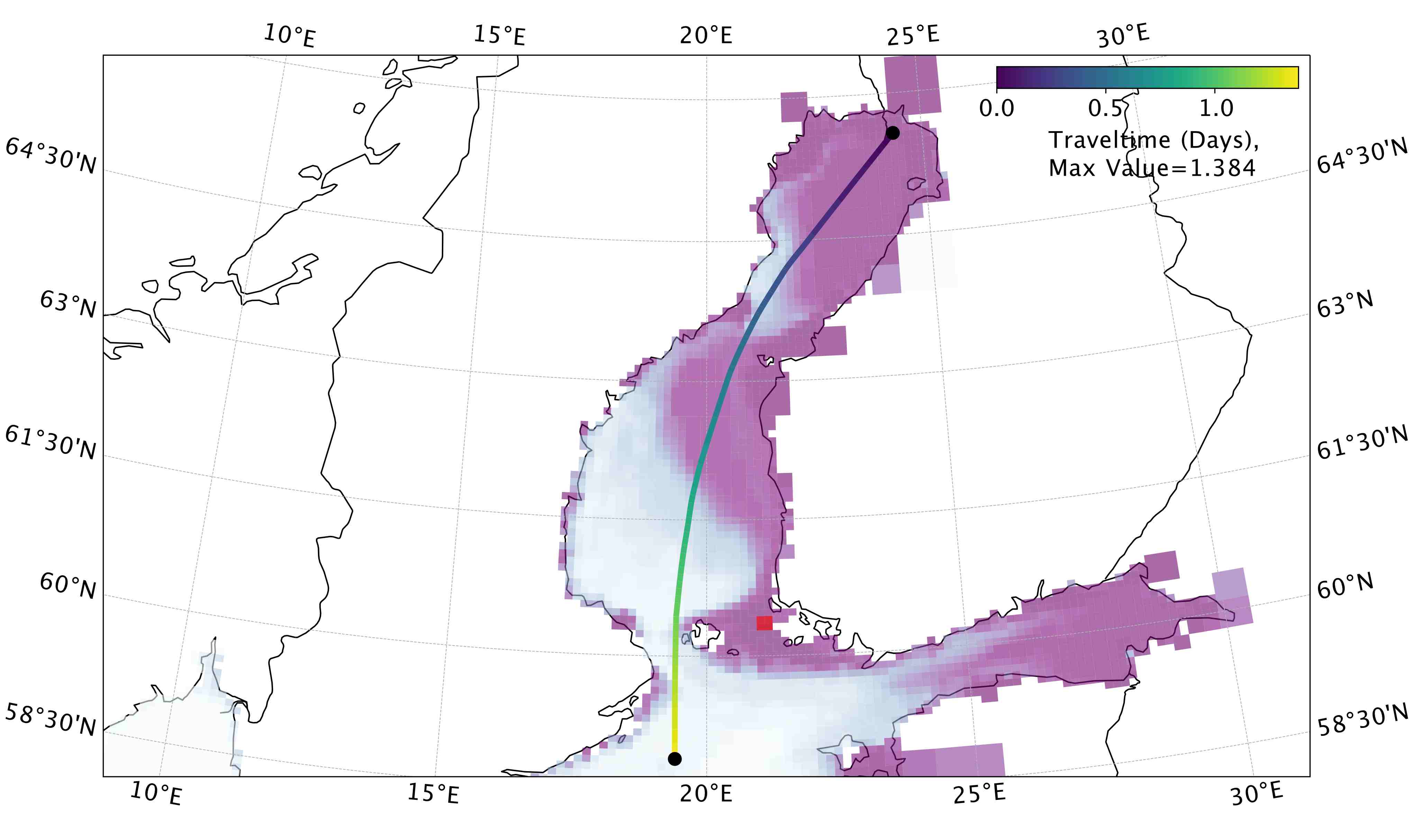}
    \caption{The route from Kemi to Stockholm in the Baltic Sea, March 2011. This can be compared with the magenta path (the experienced seafarer's path) shown in Figure 10 part (b) in~\cite{Lehtola2019}}.
    \label{fig:lehtola}
\end{figure}

Figure~\ref{fig:lehtola} shows a route planned in March 2011 for a class 1A icebreaker vessel as described in~\cite{kuuliala2017} (the month, year and vessel used for the experiments in~\cite{Lehtola2019}). The route goes from Kemi to a point off the coast near Stockholm, in the Baltic Sea, and can be compared with Figure 10(b) in~\cite{Lehtola2019}. The discrepancies between our path and the black line shown in their figure (the optimal route generated by \cite{Lehtola2019}) appear to be -- at least in part -- due to the smoothing methods used. Our smoothing method has resulted in a more efficient path without excursions due to mesh artifacts such as grid points. The magenta line in their figure, which is the path attributed to an experienced seafarer, is more similar to our path but is clearly based on additional information that our approach does not model. This additional information could include shipping lanes. Our approach could capture preference for following shipping lanes (as well as applying appropriate speed restrictions while following them) by marking cells that depart from them as inaccessible during smoothing. This will be addressed in future work.

\section{Discussion and Conclusions}
 We have presented a method for automating route-planning for a vessel operating in the complex environment of the polar oceans. As has been noted by other authors \citep{Kotovirta2009, Lehtola2019, Mishra2021}, sea ice extent (as well as other environmental features) has a significant impact on the travel times and fuel requirements of journeys. Our method provides decision support to the expert navigators, functioning in a way similar to in-car navigation systems in being able to rapidly generate, evaluate and re-plan routes in changing conditions. We have developed an approach that combines the best of discrete and continuous modelling and path-planning, resulting in a system that offers some improvements over the paths generated by previous methods and is suitably responsive for practical use in difficult conditions. The approach exploits realistic temporal abstractions and is highly flexible in terms of the data that can be ingested in the modelling phase. The resulting environment models are detailed and well-founded, and the planned routes can be explained in as much detail as required by reference to the properties of the underlying mesh.

Our future work will consider the following two priorities: managing the risk of the vessel being beset in ice when navigating in ice-prone regions; taking into account wind speed and direction, both in route-planning (understanding how wind affects the power requirements of the vessel) and in in-ice modelling (understanding how wind affects the movement of the ice, and hence the opening and closing of leads). Subsequently we will consider how route-planning can be embedded in a multi-year science cruise logistics planning method that optimises the selection of science experiments as well as the best way to travel between them. 

We have discussed how differences in seasonal SIC can drastically affect the accessibility of waypoints and hence the routes taken. Since polar sea ice characteristics are changing as a consequence of changes to the climate, historic SICs are not likely to remain a reliable proxy to current conditions. The methods developed throughout this article have been formulated to allow different types of SIC information from different datasets, and as such could use forecast SICs (\cite{andersson2021,morioka2021}) to replace, or augment, the historic SIC data. Although beyond the scope of this manuscript, our future work will develop the integration of forecast sea ice data to allow the use of short-term sea ice predictions, especially when navigating locally in ice.  

The vehicle characteristics outlined in this article use initial information about the vessel's modelled response to changing conditions. As data is collected from onboard  instruments during navigation, the information gathered about the performance of the vessel during ice breaking and open water navigation could be used to create a more bespoke modelled understanding for the response of a particular vessel to the changing conditions. Although these bespoke models would be different for each vessel, the method of training from the data would be similar, and once trained these models could be interchanged, saved and used in the route planner for path construction. In future work, as more data is made available, the SDA will be used as a example test case to leverage real time data to update and improve the model of the vessel's response to changing conditions. This will improve the precision of the travel-times and fuel requirements associated with the generated routes.  

Finally, the methods described in this paper are fully generic and can therefore be applied to route planning for any controllable marine vehicle, including autonomous underwater vehicles, surface vehicles and other shipping. In particular, the methods have been built on top of the foundation established by~\cite{Fox2021} which was developed for underwater vehicle route-planning. In future work we propose to consider route planning for mixed fleets of vehicles, with the goal of maximising science gains while minimising carbon emissions and fuel expenditure.

\section*{Acknowledgements}
This project is supported by the NERC UKRI Carbon Fund. We thank Nopi Exizidou (BAS Net Zero Transition Lead) for her role in accessing this funding, and for supplying summary SDA figures for parameterising our speed adjustment model. Code developed and described in this manuscript can be found at the Github \url{https://github.com/antarctica/PolarRoute} upon release of the publication.  The corresponding executable examples used to generated the figures in this article are also given at \url{https://github.com/antarctica/PolarRoute-PaperExamples}.

\section*{Appendix}
\appendix

\begin{table}[h!]
\centering
 \def\arraystretch{1.2}%
 \begin{tabular}{|c@{\hskip 0.5in} c c|} 
 \hline
 Parameter & Slender Hull & Blunt Hull \\ [0.5ex] 
 \hline\hline
 $k_c$ & 4.4 & 16.1 \\ 
 $b$ & -0.8267 & -1.7937 \\
 $n$ & 2 & 3 \\ [1ex] 
 \hline
 \end{tabular}
\caption{Constant parameters of the ice resistance model for different hull geometries, taken from~\cite{Li2020}.}
\label{tab:iceparams}
\end{table}


\begin{table}[!ht]
    \centering
    \begin{tabular}{|l|l|l|}
    \hline
        Waypoint Name & Latitude & Longitude \\ \hline \hline
        Falklands & -52.63472222222222 & -59.879999999999995 \\ \hline
        SouthGeorgia & -54.87916666666667 & -37.264166666666654 \\ \hline
        SouthSandwichTrench & -56.41138888888889 & -25.058611111111134 \\ \hline
        NorthPeninsula & -63.03138888888889 & -50.94 \\ \hline
        MaudRise & -66.08277777777778 & -2.8274999999999864 \\ \hline
        AmundsenSea & -69.32305555555556 & -129.69083333333333 \\ \hline
        CentralWeddellSea & -71.08472222222221 & -44.94527777777779 \\ \hline
        SouthOrkneyIslands & -60.48611111111112 & -45.30472222222221 \\ \hline
        CentralScotiaSea & -58.35805555555556 & -44.232777777777756 \\ \hline
        FilchnerTroughOverflow & -76.99277777777777 & -35.946944444444455 \\ \hline
        ArgentineSea & -43.0 & -58.13388888888892 \\ \hline
        BellingshausenSea & -70.90333333333334 & -82.8175 \\ \hline
        MargueriteBay & -68.38916666666667 & -69.24361111111114 \\ \hline
        Palmer & -64.67861111111111 & -67.13305555555553 \\ \hline
        Brunt & -74.04166666666667 & -28.28472222222223 \\ \hline
        NorthernWeddellSea & -63.44861111111111 & -34.20194444444445 \\ \hline
        NorthWestGeorgiaRise & -52.35694444444445 & -36.9377777777778 \\ \hline
        ShagRocksPassage & -52.21388888888889 & -48.155555555555566 \\ \hline
        ElephantIsland & -60.54722222222222 & -55.181388888888875 \\ \hline
        BurdwoodBank & -54.48527777777778 & -60.138888888888914 \\ \hline
        A23bottom & -62.43418611111111 & -33.26091111111111 \\ \hline
        A23top & -57.02050833333333 & -38.35846388888888 \\ \hline
        SR4top & -61.5827 & -49.756022222222214 \\ \hline
        SR4bottom & -68.41325555555557 & -29.313119444444453 \\ \hline
        OrkneyPassageEntry & -60.951438888888894 & -28.578744444444453 \\ \hline
        OrkneyPassageExit & -60.694494444444445 & -39.17623611111111 \\ \hline
        WSDWsource1 & -66.77425833333334 & -57.253008333333355 \\ \hline
        WSDWsource2 & -70.76821111111111 & -52.97460833333332 \\ \hline
        LongPathStart & -58.0 & -58.0 \\ \hline
        LongPathEnd & -58.0 & -129.0 \\ \hline
        SassiMoorings & -72.25028 & -18.62722 \\ \hline
        Halley & -75.26722 & -27.21694 \\ \hline
        AmundsenSeaInlet & -73.438 & -106.875 \\ \hline
    \end{tabular}
    \caption{List of user defined waypoint locations used throughout this work}
    \label{tab:waypoints}
\end{table}

\bibliography{main}

\end{document}